\documentclass[11pt]{article}
\usepackage{yandex-research}

\usepackage{amsmath,amsfonts,bm}

\def\eqref#1{equation~\ref{#1}}

\def\1{\bm{1}}

\DeclareMathAlphabet{\mathsfit}{\encodingdefault}{\sfdefault}{m}{sl}
\SetMathAlphabet{\mathsfit}{bold}{\encodingdefault}{\sfdefault}{bx}{n}

\usepackage{hyperref}
\usepackage{url}
\usepackage{graphicx}
\usepackage{booktabs}
\usepackage{microtype}
\hypersetup{
  pdftitle={On Data Scaling in Tabular ML: More Data, Better Generalization?},
  pdfauthor={Ivan Rubachev, Nikita Lazarev, Artem Babenko}
}

\usepackage{enumitem}
\setlist[itemize]{leftmargin=*,topsep=2pt,itemsep=5pt,parsep=0pt,partopsep=0pt}

\title{
  On Data Scaling in Tabular ML: \\
  More Data, Better Generalization?
}

\author{
Ivan Rubachev\textsuperscript{$\alpha,\beta$},\enspace
Nikita Lazarev\textsuperscript{$\beta,\alpha$},\enspace
Artem Babenko\textsuperscript{$\alpha,\beta$}}
\affiliations{\textsuperscript{$\alpha$}Yandex,\enspace
\textsuperscript{$\beta$}HSE University
}

\usepackage{placeins}
\usepackage{tabularx,float}
\begin{document}

\maketitle

\begin{abstract}

Recent work in tabular machine learning has emphasized the importance of diverse and realistic evaluation settings. The niche of large-scale (1M+) data, however, remains under-explored. To study training-data scaling, we assemble eight real-world tabular datasets with up to around ten million training examples, retaining the natural distribution shifts of the tasks and using random-split evaluation as a control. We find that GBDTs and MLP-based models show broadly similar scaling trends, with gains from both additional training data and algorithmic improvements. Tabular foundation models show more varied scaling behavior, with recent versions benefiting more consistently from larger contexts even under distribution shift. However, scaling training data under distribution shift introduces new challenges: in-distribution and out-of-distribution performance mostly improve together at smaller scales but can diverge at larger scales. We examine this through MLP capacity sweeps and comparisons of the validation distributions used for tuning and early stopping in MLPs and GBDTs. We also compare common practices of using validation data, contrasting its use for model selection in MLP pipelines with using it for in-context fitting in tabular foundation model pipelines. Under stronger shifts, MLPs selected using validation representative of the evaluation conditions generally maintain or improve performance as training data grows. In contrast, including such examples in context improves foundation-model performance, but does not prevent deterioration with increasing training size in our experiments.

\end{abstract}

\begin{paperlinks}
\paperlink[github.com/yandex-research/tabular-dl-more-data]{Code}{https://github.com/yandex-research/tabular-dl-more-data}
\end{paperlinks}

\section{Introduction}

Machine Learning (ML) on tabular data has recently grown more popular and mature. The field had its foundation model moment \citep{hollmannAccuratePredictionsSmall2025, quTabICLv2BetterFaster2026,zhangLimiX2ContextualMechanism2026,jagerTabPFN35TechnicalReport2026}, produced improvements to MLP-based DL models \citep{gorishniyTabMAdvancingTabular2025,gorishniyTabPackEfficientHyperparameter2026a,holzmullerBetterDefaultStrong2024}, and started forming sound benchmarking practices \citep{gorishniyRevisitingDeepLearning2021,ericksonTabArenaLivingBenchmark2025,talent}.

Recent work has highlighted the importance of focusing on more diverse and realistic evaluation settings in tabular ML.
One such setting concerns the presence of distribution shift between training and evaluation, which is common in many practical tabular prediction tasks \citep{gardnerBenchmarkingDistributionShift2024,rubachevTabReDAnalyzingPitfalls2024a,puruckerIIDHowGeneral2026}. Another
practically important aspect is the availability of large amounts of data in some applications where tabular ML methods are used \citep{anilFactoryFloorML2022,zouLargeScaleSearch2022,ploshkinYambda5BLargeScaleMultiModal2025}. This practical relevance, together with the role of data scaling in the success of DL in other modalities \citep{hestnessDeepLearningScaling2017,ardalaniUnderstandingScalingLaws2022,wengScalingLawsCarefully2026}, motivates a deliberate study of data scaling in tabular ML.

To study data scaling in settings relevant to real-world applications of tabular ML, we need large datasets with task-specific evaluation splits that retain natural distribution shifts, which remain scarce in
existing benchmarks.
To enable this study, we assemble a collection of eight datasets from prior work, open ML
competitions and adjacent domains, with training sizes reaching around ten million examples. Modern tabular benchmarks typically compare methods across vast collections of datasets \citep{ericksonTabArenaLivingBenchmark2025,talent} of different properties. Across these datasets, sample size varies together with the prediction task, feature space, and evaluation conditions. Comparing performance across tasks of different sizes therefore does not isolate the effect of dataset size. We instead study scaling within individual tasks, varying training size from 10\,000 to around ten million examples while keeping features and evaluation sets fixed. We use natural shifted evaluation alongside in-distribution controls and examine whether gains from additional data persist under distribution shift, and how this depends on model selection.

First, we evaluate GBDTs, MLP-based models and tabular foundation models across training sizes. Non-foundation DL models and GBDTs show similar scaling trends, with task-dependent saturation, and improvements to architectures and training procedures remain beneficial across scales. Foundation models (TabICL~v2, Kumo, TabPFN~v3 and v3.5) differ substantially, with recent versions benefiting more consistently from larger contexts.

Our analysis further shows that the benefits of additional data increasingly depend on model selection considerations. First, we perform wide sweeps of MLP parameter sizes and hyperparameter configurations, varying the validation distribution used for tuning and early stopping. In the in-distribution evaluation setting, tuning favors larger models, increasingly with the growing training set size. When using shifted validation and test splits, however, increasing model size beyond moderate parameter counts brings little or no consistent benefit, even as additional training data continues to improve performance. This growing disagreement over model capacity suggests that the choice of validation distribution becomes more consequential at larger scales. We test this directly for MLPs and GBDTs, comparing ID- and OOD-based tuning and early stopping. With OOD validation, OOD performance generally remains stable or improves as training size grows. With ID validation, additional data can instead improve ID performance while worsening OOD performance on several tasks, after a certain training set size threshold, with more pronounced performance deterioration observed in MLPs.

This leads us to investigate how validation, better aligned with the held-out test distribution, is used in different tabular modeling paradigms. Typical MLP pipelines use validation as a held-out signal for hyperparameter tuning and early stopping. In our experiments with stronger distribution shifts, OOD performance under this procedure generally remains stable or improves as training size grows. In ICL-based foundation models, using validation as a separate signal is not natively supported; a common default is to add it to the context. Our experiments show that this improves performance, however, it does not prevent OOD performance decrease as we scale the training dataset size, hinting that ICL-based foundation models may also have a hard time balancing both ID and OOD performance.

This motivates further investigation of how best to use data more aligned with the deployment/test distribution in both MLP-based models and PFNs. This raises the question of how to combine its roles in model selection and training while mitigating the negative ID--OOD relationship that can arise as in-domain data scales.

We summarize our contributions as follows:

\begin{itemize}[leftmargin=2.5em]
\item We study how training data size and natural distribution shift jointly affect tabular models, with in-distribution evaluation as a control. To this end, we assemble eight tabular datasets with up to around ten million training examples, and vary training size within each task while keeping features and evaluation sets fixed.

\item We find broadly similar scaling trends across parametric models (GBDTs and MLP-based models), with improvements to architectures and training procedures remaining beneficial across scales we consider. Tabular foundation models show substantially different scaling behavior across pretrained versions, with recent models benefiting more consistently from larger contexts, even in challenging out-of-distribution conditions.

\item We find that the validation distribution becomes increasingly important for OOD generalization as training data scales. On several datasets, additional training data improves ID performance but worsens OOD performance when MLPs are selected using ID validation, exhibiting an ``\textit{accuracy on the wrong line}''-like pattern \citep{sanyalAccuracyWrongLine2025} on \textit{real} tabular tasks with \textit{natural} shifts. LightGBM also shows sensitivity to the validation distribution, but ID-based selection mainly leads to smaller OOD gains or mild deterioration.

\item We explore the above effect further and find that validation aligned with the evaluation distribution helps MLP-based models improve as in-distribution training set size grows as it constrains model selection (including tuning and early stopping). For ICL-based foundation models, a common strategy of using the whole available data for in-context ``training'' improves the overall performance but does not prevent OOD performance decrease at larger training set scales.

\end{itemize}

\section{Related Work}

\textbf{Tabular ML Evaluation.} Modern tabular ML benchmarks typically compare models across a set of diverse tabular
prediction tasks
\citep{ericksonTabArenaLivingBenchmark2025,talent,gorishniyTabPackEfficientHyperparameter2026a}.
Recent work has extended these evaluations to tasks with temporal and grouped components with natural distribution
shifts \citep{rubachevTabReDAnalyzingPitfalls2024a,puruckerIIDHowGeneral2026}. Evaluations at multi-million-sample
scales have sometimes appeared in prior work
\citep{talent,gorishniyTabMAdvancingTabular2025,gorishniyTabPackEfficientHyperparameter2026a,priorlabsTabPFNScalingMode,richmanScalingLawsTabular2026},
though controlled within-task data scaling on diverse realistic tabular datasets has received relatively little attention. We complement the existing efforts
with a systematic study of within-task data scaling up to around ten million training examples under natural
distribution shifts, with in-distribution evaluation as a control.

\textbf{Data Scaling.} Our focus on scaling the available task-specific training data is closely related to the study of
\citet{hestnessDeepLearningScaling2017}, they observed predictable gains from additional data across language, vision,
and speech domains, with model improvements primarily shifting the scaling curves rather than changing their power-law
exponents. We observe similar trends for parametric tabular models. In deep learning recommendation models,
\citet{ardalaniUnderstandingScalingLaws2022} likewise found continued gains from additional data but limited returns to
parameter scaling in the architectures studied. We find a related contrast in tabular prediction, where the benefits of
additional capacity strongly depend on the evaluation distribution.

For tabular foundation models, multiple existing works study the scaling of the in-context learning pretraining stage
\citep{zhangLimiX2ContextualMechanism2026,GoogleresearchTabfm2026,maTabDPTScalingTabular2025}, in contrast we study the
downstream task scaling, increasing the in-context examples, which is closer in spirit to many-shot in-context learning
\citep{agarwalManyShotInContextLearning2024}. We observe substantial differences in scaling behavior across different
pretrained model versions, with recent models benefiting more consistently from larger contexts and, in some cases,
continuing to improve with up to one million context examples.

\textbf{Domain Generalization.} Understanding how models generalize beyond their training distribution is a
long-standing question in deep learning \citep{gulrajaniSearchLostDomain2020,kohWILDSBenchmarkIntheWild2021,
  malininShiftsDatasetReal2022}. One prominent observation, put forward by \citet{millerAccuracyLineStrong2021} as
``\textit{accuracy on the line},'' is that models with better in-distribution (ID) performance tend to perform better
out of distribution (OOD). In tabular ML, \citet{gardnerBenchmarkingDistributionShift2024,
  nastlCausalPredictorsGeneralize2024} report similar positive correlations. However, \citet{teneyIDOODPerformance2023}
show that ID and OOD performance can also be negatively correlated on real-world datasets. Of particular relevance to
data scaling, \citet{sanyalAccuracyWrongLine2025} show how label noise and nuisance features can produce negative
correlations, with additional training data improving ID performance while worsening OOD performance. We investigate how
this relationship depends on training-set size in real-world tabular tasks, and what it implies for model selection
across scales.

\section{Setup}
\label{sec:setup}

\subsection{Datasets}

We use eight tasks, summarized in Table~\ref{tab:dataset-properties}. For Cooking Time, Delivery ETA, Maps Routing, and
Weather from TabReD \citep{rubachevTabReDAnalyzingPitfalls2024a}, we retain the existing features and temporal splits
and use the larger training sets. We additionally prepare the following four tasks, contributing feature-construction
and split recipes for future work.

\textbf{Jane Street.} We use the public training data from the Jane Street forecasting competition,\footnote{\scriptsize
  \url{https://www.kaggle.com/competitions/jane-street-real-time-market-data-forecasting/data}} which contains
anonymized financial market observations derived from production trading systems. The task is to predict the anonymized
market response \texttt{responder\_6} from market features and historical responses. Following the competition protocol
and public solutions, we construct lag features, including responses from earlier validation and test observations. We
use the final two periods of 100 date timestamps each for validation and test, respectively, and the preceding data for
training.

\textbf{Aeolus Flight Delay.} We use the tabular component of the Aeolus dataset
\citep{xuAeolusMultistructuralFlight2025a}. The task is to predict whether a flight will depart more than 15 minutes
late. We construct calendar and route features, airport-level weather climatology, and smoothed historical delay rates
for airports, carriers, and carrier--airport pairs. Historical delay rates use only training-period flights scheduled on
earlier calendar days. We use 2016--2022 for training, 2023 for validation, and 2024 for test.

\textbf{Bank Fraud.} We use the Guardian task from Data Fusion Contest 2026.\footnote{\scriptsize
  \url{https://ods.ai/competitions/data-fusion2026-guardian}} The task is to identify fraudulent transactions using
transaction information and customer history. Following public solutions, we construct transaction features and
unlabeled customer-history features from the released logs. We use the final two consecutive two-week periods for
validation and test, respectively, and the preceding data for training. Following the post-competition analysis, for
each customer in an evaluation period, we retain all days with fraud if any are present, and sample one random day
otherwise. For training, we sample negatives at a 15\% rate. We draw ID splits from the training period with the same
class balance as the temporal splits. We use larger evaluation sets than for most other tasks to reduce noise in AP
estimates, since positives constitute only about 0.5\% of evaluation examples.

\textbf{Istella.} We use the Istella22 search relevance dataset \citep{datoIstella22DatasetBridging2022}. The task is to
estimate the relevance of a document to a search query. We treat the graded relevance labels as regression targets and
retain the 200 non-constant query-document features. We combine the official training and validation partitions for
training, and construct validation and test sets from disjoint query groups in the official test partition, whose
distribution differs from those of the training and validation partitions. Each holdout contains approximately 600\,000
query-document pairs, about 40\% of the official test partition, sampled by query group. We use these larger evaluation
sets to reduce noise in performance estimates, since they contain only 895 validation and 860 test query groups.

\subsection{Evaluation Protocol}

In this section we describe the details of our evaluation protocol, which we reuse in the next sections.

\textbf{Dataset splits}. Each task has validation and test sets under its temporal or query-based split for
out-of-distribution (OOD) evaluation, together with historical row holdouts as in-distribution (ID) controls. ID rows
are reserved before subsampling and match the OOD holdout sizes. For each dataset, validation and test sets remain fixed
across all training sizes and models. All models use the same training subsets at each size. Training subsets are nested and the sizes are doubling from 10\,000 to 5\,120\,000, followed by the largest subset which is capped at 10M for datasets significantly larger than this size, and kept for others.

\textbf{Non-foundation tabular ML models}. 
We evaluate XGBoost, LightGBM, MLP, MLP with piecewise-linear embeddings (denoted by MLP$^{\ddagger}$) \citep{gorishniyEmbeddingsNumericalFeatures2022}, MLP trained with Muon optimizer \citep{gorishniyBenchmarkingOptimizersMLPs2026}, TabM \citep{gorishniyTabMAdvancingTabular2025}, TabPack and TabPack with its variant of periodic embeddings \citep{gorishniyTabPackEfficientHyperparameter2026a} (denoted by TabPack$^{\dagger}$). We tune each method using Optuna \citep{akibaOptunaNextgenerationHyperparameter2019} separately at each dataset and training size with 100 trials, except TabPack, which selects a greedy ensemble from 64 sampled configurations.

\textbf{Tabular foundation models}. We evaluate TabICLv2 \citep{quTabICLv2BetterFaster2026}, TabPFN v3, v3.5 and its fast variant \citep{jagerTabPFN35TechnicalReport2026}, and Kumo Tabular in its small, medium, and large variants \citep{NVIDIAKumoTabular2026}. For all models we use the provided default configurations.

\textbf{Result reporting}.
We evaluate ten seeds for each model. We report means and standard deviations of test scores. For regression, we use root mean squared error (RMSE). For binary classification, we use average precision (AP), since positive classes are substantially less frequent than negatives.

\section{Empirical Training Data Scaling Results}

We start with experiments on how test performance changes as training sets grow under the benchmark's default splits with natural distribution shifts. We first report scaling trends for parametric tabular models in \autoref{sec:parametric-scaling}, and then study tabular foundation models in \autoref{sec:tabfm-scaling}.

\subsection{Parametric Tabular Models}
\label{sec:parametric-scaling}

\begin{figure}[!htbp]
  \centering
  \includegraphics[width=\linewidth]{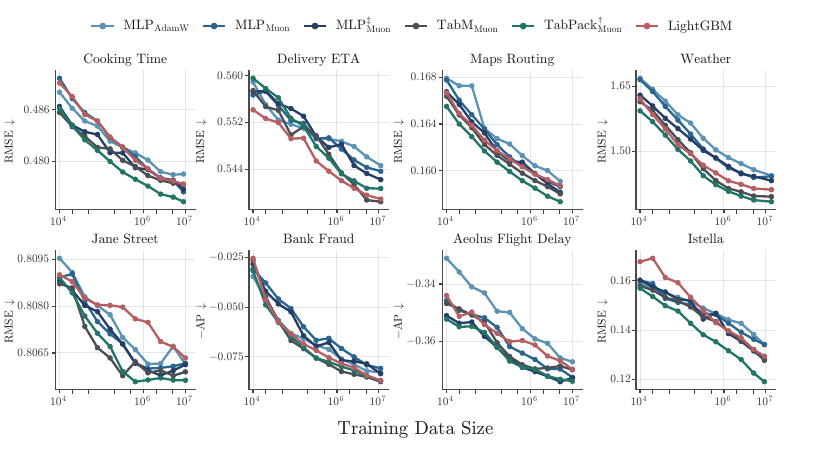}
  \caption{Performance of parametric models across training-set sizes. Horizontal axis shows the number of
    training examples on a logarithmic scale. We report RMSE for regression and negative average precision
    ($-\mathrm{AP}$) for classification, so lower is better on every plot. On most tasks, additional training data improves performance across methods, improvements to architectures and training procedures also remain beneficial across scales.}
  \label{fig:results-parametric}
\end{figure}

Results for a representative set of tabular ML models trained with increasing dataset sizes are summarized in \autoref{fig:results-parametric}, for more models see tables in the \autoref{app:full-parametric-results}. Our findings are summarized below.

\textbf{Scaling data provides significant improvements}. Overall we see that increasing the training-set size improves test performance across a broad range of parametric models. On many tasks improvements continue through the largest training sizes, approaching ten million examples. The extent of these gains depends on the task: Jane Street largely saturates around one million examples, while improvements on Weather become smaller at the largest sizes.

\textbf{Method improvements remain relevant as data scales}. Muon often improves vanilla MLP performance across training sizes, numerical feature embeddings provide further gains on several tasks. TabPack$^{\dagger}$ ranks among the top methods across most sizes considered. Although the benefits of individual changes vary across tasks and sizes and are quite noisy, additional data does not eliminate performance gaps. Improvements to models and training procedures thus complement improvements seen from data scaling at this scale.

Overall gains from additional data follow broadly similar trends across different methods within each task. This is
consistent with \citet{hestnessDeepLearningScaling2017}, who found that model improvements in other deep learning domains mainly shift the data-scaling curves while leaving their power-law exponents largely unchanged. To assess whether the similar scaling trends could be observed across methods on datasets from our evaluation we compare power-law fits with individual and shared exponents in \autoref{app:scaling-exponents} and find that individual exponents provide small improvements in fit quality (with the exception of Jane Street dataset).

\textbf{Takeaway.}
Improvements to architectures and training procedures remain beneficial as training data grows, complementing the gains from additional data. TabPack \citep{gorishniyTabPackEfficientHyperparameter2026a} offers a favorable performance--runtime trade-off among the evaluated parametric methods, combining competitive predictive performance with efficient training and model selection (\autoref{app:full-training-runtime}).

\subsection{Tabular Foundation Models}
\label{sec:tabfm-scaling}

\begin{figure}[!htbp]
  \centering
  \includegraphics[width=\linewidth]{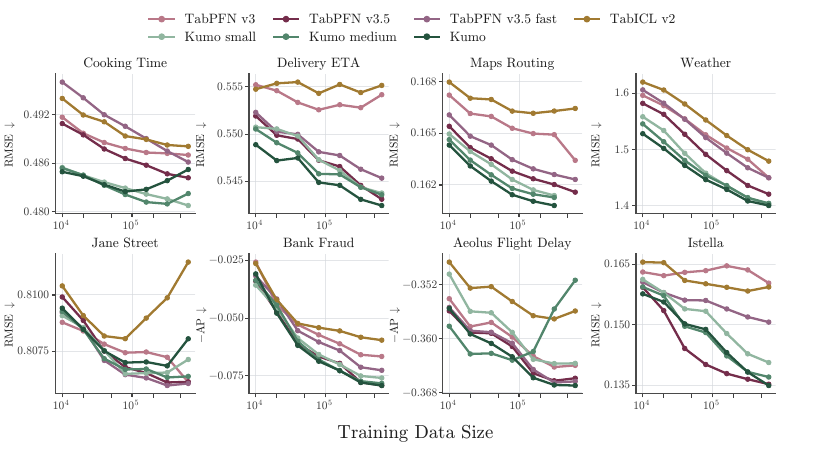}
  \caption{
    Performance of tabular foundation models across training-set sizes. The figure follows \autoref{fig:results-parametric} conventions. Tabular foundation models also improve with in-context training set scale, but the scaling trends vary significantly across different models.
  }
  \label{fig:results-pfns}
\end{figure}

Results for tabular foundation model in-context training set scaling are summarized in \autoref{fig:results-pfns}. We outline our findings below.

\textbf{Additional context improves performance under distribution shift}. Increasing the in-context
training-set size improves test performance on many tasks, with gains continuing up to around one million
examples in some cases. The extent of these gains depends on the task: curves on Jane Street, Cooking Time and
Aeolus Flight Delay can flatten or reverse at larger context sizes for some models (more on that in \autoref{sec:validation-parametric-pfns}).

\textbf{Scaling behavior differs across pretrained models}. Unlike the broadly similar scaling trends of
parametric models, foundation models differ substantially in their gains from additional context.
Kumo, TabPFN v3.5 and its fast variant show more consistent improvements with larger contexts than earlier TabPFN v3 and TabICL v2.

\textbf{Scaling behavior also differs within a model family}.
We observe different scaling trends both across Kumo model sizes and between TabPFN v3.5 and its fast variant. The small, medium and large Kumo variants differ in both parameter count and the number of synthetic tables used for pretraining \citep{NVIDIAKumoTabular2026}, yet larger variants do not consistently benefit more from additional context. TabPFN v3.5 and its fast variant also differ in the magnitude of their gains from larger contexts. How these differences relate to model capacity, pretraining and inference configurations remains an open question.

\textbf{Takeaway.}
Recent tabular foundation models benefit more consistently from larger contexts even under natural distribution shifts, providing encouraging evidence that this paradigm can scale in challenging real-world settings. However, scaling behavior varies substantially across models, and understanding the roles of model capacity, pretraining, and inference configurations remains an open question.

\section{Data Scale, Generalization and Model Selection}
\label{sec:analysis}

Sections~\ref{sec:parametric-scaling} and \ref{sec:tabfm-scaling} showed that additional training data improves out-of-distribution test performance for most models and tasks. In this section, we present a more nuanced picture: model selection, particularly the use of validation aligned with the target distribution, becomes increasingly important for OOD generalization as training data grows. We use in-distribution validation and test splits as controls for the analysis.

In \autoref{sec:analysis-setup}, we sweep MLP widths and hyperparameters across training sizes. We find that in-distribution and out-of-distribution evaluation increasingly ``disagree'' on the benefits of model capacity. In \autoref{sec:validation-at-scale}, we show that this disagreement matters for model selection, starting at around 1M examples scale in our setup. In \autoref{sec:validation-parametric-pfns}, we examine how common defaults in MLP and tabular foundation model pipelines differ in their use of target-aligned validation data, and what these choices imply for generalization as training data grows.

\subsection{Model Capacity Across Training Sizes}
\label{sec:analysis-setup}

To study how training size and model capacity jointly affect performance, we run a large sweep of MLPs over widths, hyperparameters, and training sizes. Batched packed hyperparameter ensembles introduced in \citep{gorishniyTabPackEfficientHyperparameter2026a} make such a sweep relatively inexpensive  and allow us to efficiently explore a large space of MLP models. The setup is as follows:

\begin{itemize}[leftmargin=2.5em]
\item \textbf{Models.} MLPs with three hidden blocks and 11 widths from 64 to 1936, spanning approximately $10^4$ to $10^7$ parameters.
\item \textbf{Hyperparameters.} For each width, one pack of 100 MLPs covering a $10 \times 10$ grid of learning rates and weight decays.
\item \textbf{Training.} AdamW for 100\,000 steps with batch size 1\,024, evaluating every 1\,000 steps.
\item \textbf{Selection.} For each width and training size, we select the learning rate, weight decay, and checkpoint on either the OOD or the ID validation set. Both protocols share the same training runs.
\end{itemize}

\begin{figure}[t]
  \centering
  \includegraphics[width=\linewidth]{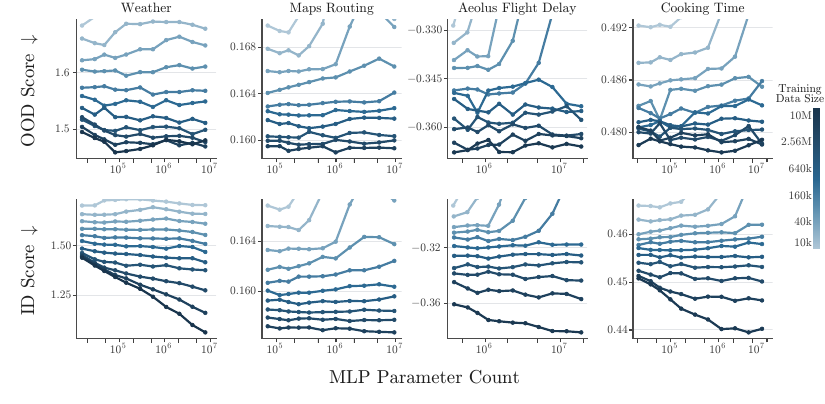}
  \caption{
  MLP width sweeps across training sizes. Top row: OOD test performance; bottom row: ID controls. The horizontal axis shows parameters per MLP on a logarithmic scale; darker curves indicate larger training sets. Learning rate, weight decay, and checkpoint are selected on the corresponding validation split. We report RMSE for all datasets except Aeolus Flight Delay, where we report negative average precision ($-\mathrm{AP}$). Lower is better. Gains from increasing model size are limited under distribution shift, but become more pronounced with larger training sets
  under ID evaluation   }
  \label{fig:sweep-model-size}
\end{figure}

In total, we train 95\,700 MLPs across eight datasets. Beyond the selected models, the stored trajectories give an overview of learning dynamics across training sizes and capacities, and allow for experimenting with different model and hyperparameter selection strategies post hoc. Results of the sweeps are summarized in \autoref{fig:sweep-model-size}. Results on all datasets are in \autoref{app:full-model-size-sweeps}. Below we provide our key takeaways.

\textbf{ID and OOD evaluation ``disagree'' on model capacity at scale}.
On OOD test sets, more training data improves performance. But, larger models bring little or no consistent gain (one can achieve best performance with around 100k parameters, on most datasets). In contrast, on in-distribution controls, larger models improve performance on almost all datasets, and these gains grow with training size. \citet{ardalaniUnderstandingScalingLaws2022} similarly report stronger gains from data than from parameters in the MLP components of DLRMs. In our setting, the benefit of capacity depends on the evaluation distribution. As training size grows, ID and OOD validation therefore increasingly favor different models. We examine the implications of this  phenomenon in the next section.

\subsection{Validation Distribution at Scale}
\label{sec:validation-at-scale}

If ID and OOD validation favor different models, the choice of validation distribution should increasingly affect OOD performance as training data scales. We test this hypothesis on MLPs and LightGBM.

\begin{figure}[!htbp]
  \centering
  \includegraphics[width=\linewidth]{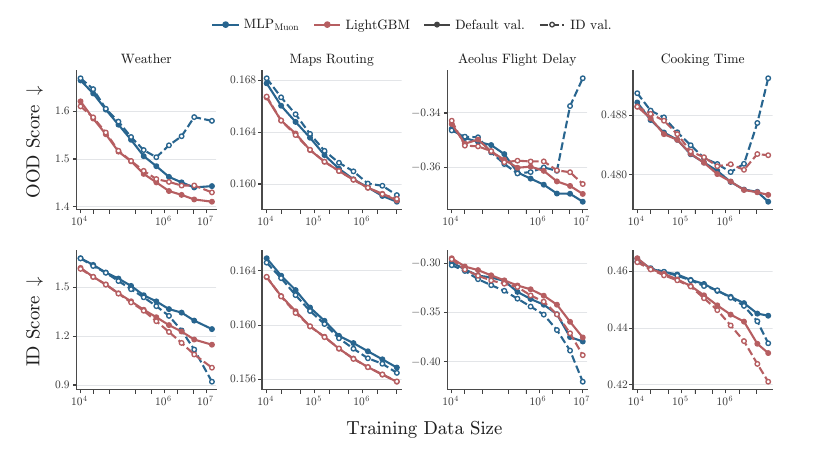}
  \caption{
    Validation distribution affects ID and OOD scaling in MLP with Muon and LightGBM. Solid curves with filled circles use OOD validation (``Default val.''); dashed curves with open circles use ID validation. We report RMSE for all datasets except Aeolus Flight Delay, where we report negative average precision ($-\mathrm{AP}$). Lower is better in all panels. At larger training sizes, ID validation can improve ID performance while OOD performance deteriorates; conversely, OOD validation can improve OOD performance at the cost of ID performance.
  }
  \label{fig:tabpack-vs-iid}
\end{figure}

We use the Muon optimizer for MLPs. We run each model using either the default OOD validation set or the ID validation set for hyperparameter tuning and early stopping, keeping training subsets and tuning budgets fixed. We evaluate both protocols on OOD test sets, with ID evaluation as a control. Results are summarized in \autoref{fig:tabpack-vs-iid}; results on all eight datasets are provided in \autoref{app:parametric-validation-splits}.

\textbf{Accuracy on the wrong line.}
With OOD-based selection, OOD performance generally improves or saturates as training size grows. At smaller training sizes, switching to ID validation usually makes little difference to OOD performance. At larger sizes, however, ID-based selection can lead to OOD deterioration despite continued improvements in ID performance. This divergence is particularly pronounced for MLPs on Weather, Aeolus Flight Delay, and Cooking Time, consistent with previously reported inverse ID--OOD relationships \citep{teneyIDOODPerformance2023,sanyalAccuracyWrongLine2025}, but observed on real tabular tasks with natural distribution shifts here. For LightGBM, ID-based selection often leads to smaller OOD gains or milder deterioration.

\textbf{Takeaway.}
Our results suggest that validation aligned with the target conditions becomes increasingly important for OOD generalization as training data grows. Prior work has examined model selection under shift and alternative validation protocols \citep{gulrajaniSearchLostDomain2020,caiUnderstandingLimitsDeep2025}. How best to achieve this alignment in practice remains an open, task-dependent question for researchers and practitioners; we highlight that the consequences of these choices can become more pronounced at larger training sizes.

\subsection{Use of Validation in MLP and Tabular Foundation Model Pipelines}
\label{sec:validation-parametric-pfns}

Previous section results showed that, at scale, validation aligned with the test distribution matters for parametric models. The two major modeling paradigms for tabular ML nowadays are parametric models (MLPs in this section) and tabular foundation models. Typical pipelines for these two paradigms differ in how they use validation data. MLP-based models often utilize validation as a held-out signal for model selection (through model tuning and early-stopping). Tabular foundation models do not have a separate validation concept and the default strategy is to fit on all the available data.

\begin{figure}[!htbp]
  \centering
  \includegraphics[width=\linewidth]{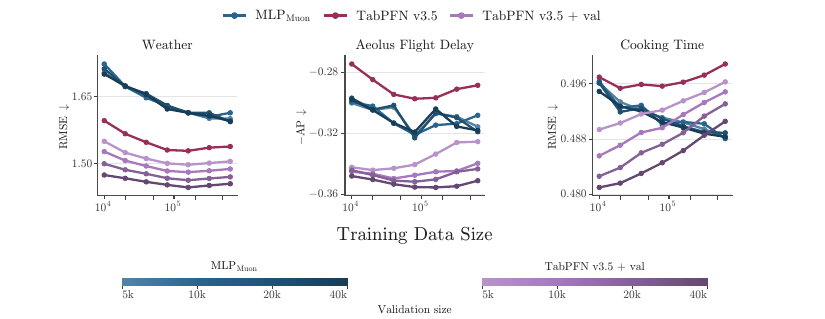}
  \caption{
    Validation as context vs.\ as a selection signal on restricted training pools of Weather, Aeolus Flight Delay, and Cooking Time. Shades of blue (MLP) and purple (TabPFN v3.5 with appended validation) indicate validation sizes from 5\,000 to 40\,000. Conventions follow \autoref{fig:results-parametric}. TabPFN benefits from additional validation examples, but adding more
training data can hurt performance. The MLP benefits more consistently
from additional training data, but gains little from larger validation sets.
  }
  \label{fig:validation-pfns-vs-parametric}
\end{figure}

We try to compare the two strategies and modeling paradigms in this section.
To imitate the effect we observed in the previous sections at sub-1M scale we manually increase the level of distribution shift between training and evaluation, intuitively this should move the ID--OOD tradeoff to smaller scale datasets. We achieve this differently by taking different subsets of the training data on Weather, Aeolus Flight Delay and Cooking Time datasets that are more OOD compared to the validation and test sets (interpretable strategies are described in \autoref{app:validation-data}). We then repeat the protocol of data scaling, but additionally we also vary the representative validation set size and evaluate two models: MLP and TabPFNv3.5.
Results are summarized in \autoref{fig:validation-pfns-vs-parametric}. Our key takeaways are as follows:

\textbf{Validation as additional context.}
Using representative validation examples in-context yields large gains at small training sizes, and larger validation sets provide increasingly larger gains. This is expected as using more relevant data for training should improve performance.

\textbf{Validation as a model-selection signal.}
However, we see on all three datasets that using all the test-aligned validation examples in ICL does not save tabular foundation models from the ``accuracy on the wrong line'' like behaviour and, with the amount of in-distribution data increasing, out-of-distribution performance starts to degrade. MLP, which uses validation for model selection, is generally more stable, and even improves as training size grows at this scale, and this does not really depend on the validation set size.

\textbf{Takeaway}. 
These two observations together illustrate that the optimal strategy of target aligned validation is not yet settled in both paradigms. MLP-like non-foundation models could benefit from procedures like refitting, and foundation models could benefit from having a more explicit way to control model selection, or specify the ``shift direction'' with some hold-out validation.

\section{Limitations}

Our work is a controlled study rather than a benchmark. In all the experiments we keep validation and test sets fixed across training sizes for comparability, which may be less representative of real-world practice at small training sizes, but provides cleaner signal for studying the effects of scaling training data. 

We do not directly compare parametric models and tabular foundation models. The two paradigms currently operate at different scales (10M scale is handled relatively easily for non-foundation models, but 1M is the current limit for foundation models), running foundation models on ten million examples requires additional techniques such as context subsampling or retrieval, which could potentially introduce confounding effects. Furthermore, as \autoref{sec:analysis} shows, a proper comparison also depends on how each paradigm uses validation data, and there is no settled answer to this yet. We also use eight datasets, the representative minimum we considered sufficient for studying within-task scaling.

\section{Conclusion}

Large training sets are common in practical and important tabular applications but they were rarely studied in tabular ML. We assembled a representative set of datasets and studied this regime. We find that progress that the tabular ML community has made holds up to scaling data. Non-foundation model architectural improvements complement improvements from scaling training data, foundation models continue to improve, and also show promising scaling trends, even in challenging realistic conditions. 

However, our analysis showed that increasing training data scale, in the presence of distribution shifts introduces new challenges. At small scales, ID and OOD performance mostly improve together. At larger scales, they can diverge, making OOD generalization a more important concern. This raises questions we find worth pursuing in future work. How should target-aligned data be split between fitting and model selection as in-domain data grows? Can foundation models be given an effective mechanism to treat validation examples differently from training examples? How do model-level generalization interventions interact with hyperparameter tuning at scale, are there new generalization inducing methods to be discovered that are particularly interesting at scale.

\subsection*{Reproducibility statement}

We supply the minimal code needed for dataset preprocessing  and running the experiments in supplementary materials. Dataset sources, feature construction, and splits are described in \autoref{sec:setup}. Hyperparameter spaces, tuning protocols, and training details are described in \autoref{app:preprocessing-tuning}.

\bibliography{references}
\bibliographystyle{unsrtnat}

\clearpage
\appendix
\raggedbottom

\section{Shared scaling exponents across methods}
\label{app:scaling-exponents}

We compare method-specific and dataset-shared power-law exponents using
71 primary-test learning curves from nine parametric methods on eight datasets.
Each point averages ten seeds; curves require at least five completed sizes
$N\ge10^4$. We fit
\begin{equation}
 E_{d,m}(N)=a_{d,m}(N/10^4)^{-b}+c_{d,m},\qquad
 b=b_{d,m}\ \text{or}\ b_d,
 \label{eq:shared-scaling-law}
\end{equation}
where $E$ is RMSE or $1-\mathrm{AP}$. Amplitudes and asymptotes remain
method-specific; a shared exponent is estimated separately for each dataset.

We minimize the mean squared residual for each curve. For the shared fit,
we normalize each curve by its error at the smallest training size, then sum
the mean squared residuals across methods. The bounds are $a,c\ge0$,
$10^{-4}\le b\le3$, and $c\le1$ for classification.


\begin{table}[!ht]
\centering
\small
\begin{tabular}{lrrrr}
\toprule
Dataset & $b_d$ & $R^2_{\mathrm{ind}}$ & $R^2_{\mathrm{shared}}$ & $\Delta R^2$ \\
\midrule
Cooking Time        & 0.1625 & 0.9938 & 0.9931 & 0.0006 \\
Delivery ETA        & 0.0055 & 0.9727 & 0.9673 & 0.0054 \\
Maps Routing        & 0.1811 & 0.9935 & 0.9929 & 0.0006 \\
Weather             & 0.1843 & 0.9921 & 0.9904 & 0.0017 \\
Jane Street         & 0.3403 & 0.9312 & 0.9024 & 0.0288 \\
Bank Fraud          & 0.3373 & 0.9930 & 0.9849 & 0.0082 \\
Aeolus Flight Delay & 0.0985 & 0.9663 & 0.9632 & 0.0031 \\
Istella             & 0.0401 & 0.9773 & 0.9752 & 0.0021 \\
\bottomrule
\end{tabular}
\caption{Dataset-shared exponents and mean in-sample $R^2$ over methods.
$\Delta R^2=R^2_{\mathrm{ind}}-R^2_{\mathrm{shared}}$, computed before rounding.}
\label{tab:shared-scaling-exponents}
\end{table}

Sharing the exponent reduces mean in-sample $R^2$ by less than $0.01$ on
7 of eight datasets (Table~\ref{tab:shared-scaling-exponents}).
Jane Street is the main exception.
This supports similar scaling across methods within most datasets over the
observed size range, while allowing different error levels and asymptotes.

\FloatBarrier
\section{Dataset construction for the validation-data experiment}
\label{app:validation-data}

To study adaptation to temporal out-of-distribution (OOD) data, we restrict
which historical periods contribute training examples while keeping the
original temporal validation and test periods:

\begin{itemize}
\item \textbf{Weather: the latest 10\% of training rows.}
We keep the latest 10\% of the original training rows in chronological
order, covering the late part of spring~2023, with validation in June and
testing in July. Restricting
seasonal coverage makes the model transfer from a narrow recent window to
a later part of the seasonal cycle. Additional training examples from this
window do not restore the missing seasonal coverage.

\item \textbf{Aeolus Flight Delay: March--December 2020.}
From the benchmark's existing training pool, we retain only flights with
scheduled departures from March~1 through December~31, 2020. Validation
uses 2023 and testing uses 2024. This isolates the pandemic-era period and
excludes the more recent 2021--2022 training data, increasing temporal
separation and testing transfer across periods with potentially different
traffic and delay patterns. The restriction applies to training rows;
the original historical-context features are retained.

\item \textbf{Cooking Time: the earliest 25\% of training rows.}
We keep the earliest 25\% of the original training rows in chronological
order. Validation remains December~21--27, 2023, and testing starts
on December~28. Removing the later training rows widens the gap to
evaluation, exposing the model to changes in the order population and
cooking-time relationships that may occur over time. Increasing the
training size still draws examples only from the earlier period.
\end{itemize}

\FloatBarrier
\section{Hyperparameter selection and evaluation runtime}
\label{app:full-training-runtime}

Table~\ref{tab:full-training-time} reports hyperparameter selection plus
single-seed training and evaluation time on the full training splits of the
main benchmark. Each value adds the recorded selection timer to the mean
individual evaluation time over ten random seeds; $\pm$ denotes the sample
standard deviation of evaluation time across seeds. Neural-model runs use
a single NVIDIA A100 GPU with 80 GB of memory. For TabPack, the reported
cost includes joint pack training and validation-based ensemble selection,
without a separate hyperparameter-tuning phase of independent training runs.

Our scaling results show that parametric models benefit from additional data
with broadly similar trends within each task, while differences in predictive
performance remain relevant across training sizes. When additional data
provides steady gains without widening the performance gap between methods,
computational efficiency becomes an important practical consideration.
In this setting, TabPack is an attractive option: it remains competitive
with the strongest tuned baselines at training scales of order $10^7$
examples, while combining training and model selection within one pack.
The substantially lower recorded runtimes in
Table~\ref{tab:full-training-time} support this efficiency advantage.

\begin{table}[!ht]
\centering
\begingroup
\small
\setlength{\tabcolsep}{4pt}
\renewcommand{\arraystretch}{1.1}
\resizebox{\linewidth}{!}{%
\begin{tabular}{@{}lrrrrrrr@{}}
\toprule
Method & Cooking & Delivery & Weather & Jane Street & Bank Fraud & Aeolus & Istella \\
\midrule
XGBoost & $24.47 \pm 0.03$ & $34.29 \pm 0.21$ & $29.37 \pm 0.01$ & $31.26 \pm 0.14$ & $51.32 \pm 0.25$ & $7.72 \pm 0.01$ & $34.24 \pm 0.09$ \\
LightGBM & $55.63 \pm 0.07$ & $70.77 \pm 0.06$ & $130.24 \pm 0.08$ & $145.78 \pm 0.76$ & $66.97 \pm 0.43$ & $127.03 \pm 0.73$ & $46.28 \pm 0.07$ \\
$\mathrm{MLP}_{\mathrm{AdamW}}$ & $3.73 \pm 0.01$ & $9.95 \pm 0.07$ & $7.27 \pm 0.04$ & $3.39 \pm 0.01$ & $7.62 \pm 0.04$ & $7.99 \pm 0.07$ & $14.03 \pm 0.13$ \\
$\mathrm{MLP}_{\mathrm{Muon}}$ & $10.55 \pm 0.06$ & $14.44 \pm 0.11$ & $10.76 \pm 0.04$ & $6.67 \pm 0.01$ & $13.30 \pm 0.02$ & $19.58 \pm 0.08$ & $14.81 \pm 0.11$ \\
$\mathrm{MLP}^{\ddagger}_{\mathrm{AdamW}}$ & $15.46 \pm 0.17$ & $18.99 \pm 0.12$ & $15.46 \pm 0.05$ & $7.96 \pm 0.03$ & $14.65 \pm 0.04$ & $14.76 \pm 0.18$ & $21.57 \pm 0.22$ \\
$\mathrm{MLP}^{\ddagger}_{\mathrm{Muon}}$ & $20.36 \pm 0.11$ & $28.41 \pm 0.40$ & $24.29 \pm 0.11$ & $9.98 \pm 0.02$ & $17.20 \pm 0.06$ & $31.74 \pm 0.17$ & $34.53 \pm 0.24$ \\
$\mathrm{TabM}_{\mathrm{Muon}}$ & $22.80 \pm 0.26$ & $37.78 \pm 0.18$ & $42.12 \pm 0.29$ & $17.07 \pm 0.07$ & $40.26 \pm 0.41$ & $52.51 \pm 0.22$ & $67.06 \pm 0.21$ \\
\midrule
$\boldsymbol{\mathrm{TabPack}_{\mathrm{Muon}}}$ & $\boldsymbol{1.09 \pm 0.05}$ & $\boldsymbol{1.74 \pm 0.08}$ & $\boldsymbol{1.60 \pm 0.07}$ & $\boldsymbol{1.37 \pm 0.04}$ & $\boldsymbol{1.67 \pm 0.04}$ & $\boldsymbol{3.03 \pm 0.07}$ & $\boldsymbol{1.35 \pm 0.05}$ \\
$\boldsymbol{\mathrm{TabPack}^{\dagger}_{\mathrm{Muon}}}$ & $\boldsymbol{6.06 \pm 0.05}$ & $\boldsymbol{8.72 \pm 0.07}$ & $\boldsymbol{4.92 \pm 0.05}$ & $\boldsymbol{3.58 \pm 0.08}$ & $\boldsymbol{4.08 \pm 0.09}$ & $\boldsymbol{3.97 \pm 0.11}$ & $\boldsymbol{5.89 \pm 0.08}$ \\
\bottomrule
\end{tabular}%
}
\endgroup
\caption{Hyperparameter selection plus single-seed training and evaluation
time on the full training splits, in hours; $\pm$ denotes the sample standard
deviation of evaluation time over ten seeds. TabPack's cost includes pack
training and ensemble selection. Together with the broadly similar scaling
trends and competitive predictive performance, these runtimes highlight
TabPack as an efficient alternative to separately tuned models at large
training sizes. TabPack rows are shown in bold.}
\label{tab:full-training-time}
\end{table}
\FloatBarrier
\section{Preprocessing and hyperparameter tuning}
\label{app:preprocessing-tuning}

\subsection{Dataset Metadata}

\begin{table}[H]
\centering
\small
\setlength{\tabcolsep}{1pt}
\setlength{\belowcaptionskip}{10pt}
\caption{Dataset statistics (rounded).
Training counts are in millions; validation and test counts
are in thousands, per evaluation condition.
The first four datasets are from TabReD.
OOD splits are temporal except for Istella (query-based).
R: regression; B: binary classification.}
\label{tab:dataset-properties}
\setlength{\tabcolsep}{4pt}
\renewcommand{\arraystretch}{0.8}
\begin{tabular}{@{}lrrrrc@{}}
\toprule
Dataset & Train (M) & Val. (K) & Test (K) & Features & Task \\
\midrule
Maps Routing        &  5.12 &  60 &  60 & 986 & R \\
Cooking Time        &  9.00 &  51 &  42 & 192 & R \\
Delivery ETA        & 11.08 &  34 &  37 & 223 & R \\
Weather             & 13.54 &  42 &  41 & 103 & R \\
\midrule
Istella             &  9.21 & 600 & 600 & 200 & R \\
Jane Street         & 10.00 &  93 &  93 &  94 & R \\
Aeolus Flight Delay & 10.00 & 100 & 100 &  91 & B \\
Bank Fraud          & 10.97 & 591 & 590 & 119 & B \\
\bottomrule
\end{tabular}
\end{table}

\subsection{Experimental setup}

\paragraph{Data preprocessing.}
Our default numerical preprocessing for MLP, MLP-PLE, TabM and TabPack is
quantile normalization to a normal distribution
\citep{gorishniyRevisitingDeepLearning2021,gorishniyTabMAdvancingTabular2025}.
Missing numerical values are filled with zero, constant columns are removed,
and binary features are treated as categorical.
MLP, MLP-PLE, TabM and TabPack use one-hot categorical encoding; tree models
use native categorical handling without additional numerical normalization.
Except for Kumo, regression targets are standardized using training statistics;
predictions are transformed back before evaluation. Foundation-model
preprocessing is described below.

\paragraph{Training neural networks.}
We minimize mean squared error for regression and binary cross-entropy for
classification, using constant learning rates. MLP/MLP-PLE and TabM use
gradient clipping at norm~1 and patience parameter~16. Batch size depends on
the training set size $N$ as follows.

\par\addvspace{4pt}\noindent
\begin{minipage}{\linewidth}
\centering\small
\setlength{\tabcolsep}{7pt}
\renewcommand{\arraystretch}{1.05}
\begin{tabularx}{\linewidth}{@{}Xr@{}}
\toprule
Training set size $N$ & Batch size \\
\midrule
$N\le20\,000$ & 256 \\
$20\,000<N<160\,000$ & 512 \\
$160\,000\le N<1\,280\,000$ & 1024 \\
$1\,280\,000\le N<5\,120\,000$ & 2048 \\
$N\ge5\,120\,000$ & 4096 \\
\bottomrule
\end{tabularx}
\end{minipage}\par

\paragraph{Hyperparameter tuning and evaluation.}
We use 100 TPE trials per dataset and training size for MLP, MLP-PLE,
TabM, XGBoost and LightGBM. TabPack does not require a separate
hyperparameter-tuning phase; its pack-based model selection is described
below. We select minimum validation RMSE or maximum AP.
Selected configurations are retrained for evaluation. Hyperparameters and
checkpoints are selected using the validation split. The tables below give
the default search spaces.

\subsubsection{Tabular foundation models}
\label{app:foundation-model-setup}

We use pretrained models without task-specific fine-tuning or hyperparameter
search. In the main scaling experiments, the context contains only the sampled
training subset. Validation examples are appended only in the experiment
of \autoref{sec:validation-parametric-pfns}. All three model families receive
numerical features without additional normalization and ordinal-encoded
categories, retaining their internal preprocessing.

\paragraph{TabPFN.}
We evaluate v3, v3.5 and the v3.5 fast variant
\citep{jagerTabPFN35TechnicalReport2026} with their provided pretrained weights
and default estimator configurations. The v3.5 variants use the September~2026
standard and fast checkpoints, respectively. We explicitly identify categorical
columns to the estimator. Regression targets are standardized using training
statistics, and predictions are transformed back before evaluation.

\paragraph{TabICL.}
We evaluate TabICL v2 \citep{quTabICLv2BetterFaster2026} with its provided
pretrained weights and default ensemble and preprocessing settings. Regression
targets and predictions use the same standardization procedure as TabPFN.

\paragraph{Kumo.}
We evaluate the small, medium and large Kumo Tabular variants
\citep{NVIDIAKumoTabular2026}, pretrained on approximately 35, 71 and 137 million
artificial tables, respectively. Ordinal-encoded categorical features are
passed as categorical variables. Kumo handles regression-target scaling
internally, and we average the predicted quantiles to obtain point predictions.
We use eight estimators and query batches of 1\,024 rows, with mixed-precision
inference.

\par\addvspace{5pt}\noindent
\begin{minipage}{\linewidth}
\subsection{MLP and MLP-PLE}

Both variants use the same backbone space. MLP-PLE adds quantile-based
piecewise-linear embeddings \citep{gorishniyEmbeddingsNumericalFeatures2022,gorishniyTabMAdvancingTabular2025},
with training-set bin boundaries, a learned projection and no embedding
activation. Either AdamW or Muon is used.

\par\addvspace{4pt}\noindent
\begin{minipage}{\linewidth}
\centering\small
\setlength{\tabcolsep}{7pt}
\renewcommand{\arraystretch}{1.05}
\begin{tabularx}{\linewidth}{@{}p{0.38\linewidth}X@{}}
\toprule
Parameter & Distribution or value \\
\midrule
Number of blocks & $\mathrm{UniformInt}[1,6]$ \\
Hidden width & $\mathrm{UniformInt}[64,1024]$, step 16 \\
Dropout & $\{0,\mathrm{Uniform}[0,0.5]\}$ \\
\midrule
Embedding width (PLE only) & $\mathrm{UniformInt}[8,32]$, step 4 \\
Number of bins (PLE only) & $\mathrm{UniformInt}[2,128]$ \\
\bottomrule
\end{tabularx}
\end{minipage}\par
\end{minipage}\par

\par\addvspace{5pt}\noindent
\begin{minipage}{\linewidth}
\subsection{TabM}

TabM uses $k=16$ ensemble members with independent training batches and
Muon optimization.

\par\addvspace{4pt}\noindent
\begin{minipage}{\linewidth}
\centering\small
\setlength{\tabcolsep}{7pt}
\renewcommand{\arraystretch}{1.05}
\begin{tabularx}{\linewidth}{@{}p{0.38\linewidth}X@{}}
\toprule
Parameter & Distribution or value \\
\midrule
Ensemble size $k$ & 16 \\
Number of blocks & $\mathrm{UniformInt}[1,5]$ \\
Hidden width & $\mathrm{UniformInt}[64,1024]$, step 16 \\
Dropout & $\{0,\mathrm{Uniform}[0,0.5]\}$ \\
\bottomrule
\end{tabularx}
\end{minipage}\par
\end{minipage}\par

\par\addvspace{5pt}\noindent
\begin{minipage}{\linewidth}
\subsection{TabPack}

TabPack uses 64 independently parameterized ReLU networks with Muon.
Configurations are sampled by random search. The cosine variant
TabPack$^{\dagger}$ concatenates each numerical input with $d-1$ learned cosine
features and applies an elementwise affine transform, where $d$ is the
embedding width.

\par\addvspace{4pt}\noindent
\begin{minipage}{\linewidth}
\centering\small
\setlength{\tabcolsep}{7pt}
\renewcommand{\arraystretch}{1.05}
\begin{tabularx}{\linewidth}{@{}p{0.38\linewidth}X@{}}
\toprule
Parameter & Distribution or value \\
\midrule
Number of blocks & $\mathrm{UniformInt}[1,4]$; cosine: $\mathrm{UniformInt}[1,3]$ \\
Hidden width & 384 \\
Dropout & $\{0,\mathrm{Uniform}[0,0.5]\}$ \\
\midrule
Embedding width & $\mathrm{UniformInt}[8,32]$, step 4 \\
Initialization scale & $\mathrm{LogUniform}[0.01,10]$ \\
\midrule
Sampled configurations & 64 \\
Maximum ensemble size & 32 \\
\bottomrule
\end{tabularx}
\end{minipage}\par
\end{minipage}\par

\par\addvspace{5pt}\noindent
\begin{minipage}{\linewidth}
\subsection{Optimizers}

Optimizer and architecture parameters are tuned jointly. AdamW is used
alone for the AdamW variants and as the auxiliary optimizer for Muon.
Below, (A) denotes MLP/MLP-PLE and (B) denotes TabM/TabPack, including cosine.
The dropout choice $\{0,\mathrm{Uniform}[0,0.5]\}$ is categorical,
with equal probabilities under random search.

\par\addvspace{4pt}\noindent
\begin{minipage}{\linewidth}
\centering\small
{AdamW and auxiliary AdamW\par}\vspace{3pt}
\setlength{\tabcolsep}{7pt}
\renewcommand{\arraystretch}{1.05}
\begin{tabularx}{\linewidth}{@{}p{0.38\linewidth}X@{}}
\toprule
Parameter & Distribution \\
\midrule
Learning rate & (A) $\mathrm{LogUniform}[3\cdot10^{-5},10^{-3}]$ \\
 & (B) $\mathrm{LogUniform}[10^{-4},0.005]$ \\
Weight decay & (A) $\mathrm{LogUniform}[0.005,5]$ \\
 & (B) $\mathrm{LogUniform}[0.001,1]$ \\
\bottomrule
\end{tabularx}
\par\vspace{7pt}
{Muon\par}\vspace{3pt}
\begin{tabularx}{\linewidth}{@{}p{0.38\linewidth}X@{}}
\toprule
Parameter & Distribution \\
\midrule
Muon learning rate & (A) $\mathrm{LogUniform}[10^{-4},0.03]$ \\
 & (B) $\mathrm{LogUniform}[0.001,0.1]$ \\
\bottomrule
\end{tabularx}
\par\vspace{3pt}\footnotesize
Muon uses the auxiliary AdamW rate above; its weight decay follows the same ranges.
\end{minipage}\par
\end{minipage}\par

\par\addvspace{5pt}\noindent
\begin{minipage}{\linewidth}
\subsection{Gradient-boosted trees}

XGBoost and LightGBM use at most 10,000 trees and early stopping after
300 rounds without validation improvement.

\par\addvspace{4pt}\noindent
\begin{minipage}{\linewidth}
\centering\small
{XGBoost\par}\vspace{3pt}
\setlength{\tabcolsep}{7pt}
\renewcommand{\arraystretch}{1.05}
\begin{tabularx}{\linewidth}{@{}p{0.38\linewidth}X@{}}
\toprule
Parameter & Distribution or value \\
\midrule
Maximum depth & $\mathrm{UniformInt}[3,10]$ \\
Minimum child weight & $\mathrm{LogUniform}[10^{-4},100]$ \\
Row subsampling & $\mathrm{Uniform}[0.5,1]$ \\
Feature subsampling & $\mathrm{Uniform}[0.5,1]$ \\
Learning rate & $\mathrm{LogUniform}[0.001,1]$ \\
Split penalty $\gamma$ & $\{0,\mathrm{LogUniform}[0.001,100]\}$ \\
$\ell_1$ and $\ell_2$ regularization & $\{0,\mathrm{LogUniform}[10^{-4},5]\}$, separately \\
Growth policy & $\{\text{depthwise},\text{lossguide}\}$ \\
\bottomrule
\end{tabularx}
\end{minipage}\par
\end{minipage}\par

\par\addvspace{7pt}\noindent
\begin{minipage}{\linewidth}
\centering\small
{LightGBM\par}\vspace{3pt}
\setlength{\tabcolsep}{7pt}
\renewcommand{\arraystretch}{1.05}
\begin{tabularx}{\linewidth}{@{}p{0.38\linewidth}X@{}}
\toprule
Parameter & Distribution or value \\
\midrule
Number of leaves & $\mathrm{UniformInt}[2,256]$ \\
Minimum child samples & $\mathrm{UniformInt}[5,256]$ \\
Minimum leaf Hessian sum & 0.001 \\
Row subsampling & $\mathrm{Uniform}[0.5,1]$, every round \\
Feature subsampling & $\mathrm{Uniform}[0.5,1]$ \\
Learning rate & $\mathrm{LogUniform}[0.005,0.1]$ \\
$\ell_2$ regularization & $\{0,\mathrm{LogUniform}[0.1,10]\}$ \\
Extra trees & $\{\text{false},\text{true}\}$ \\
\bottomrule
\end{tabularx}
\par\vspace{3pt}\footnotesize
For both tree models, zero-or-log-uniform parameters use a categorical switch.
\end{minipage}\par

\FloatBarrier
\section{Full model-size sweeps}
\label{app:full-model-size-sweeps}

\begin{figure}[!ht]
  \centering
  \includegraphics[width=\linewidth]{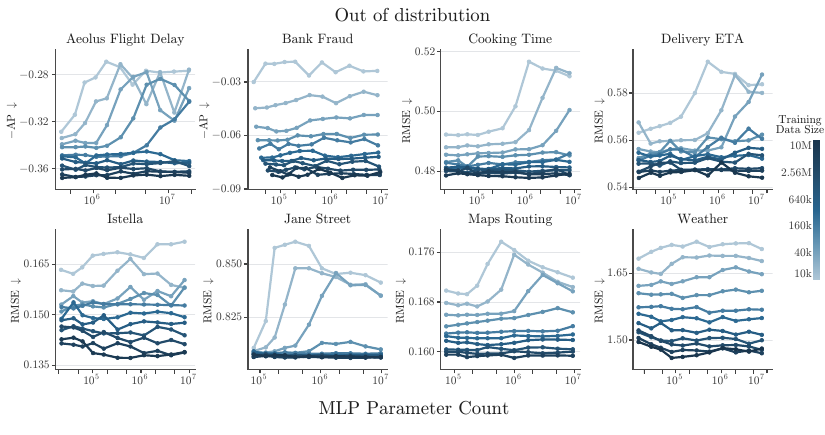}
  \caption{Default protocol: temporal validation and test splits, or query for
  Istella. Points show seed means; shaded bands show one standard deviation
  where multiple seeds are available.}
  \label{fig:full-sweep-default}
\end{figure}

\begin{figure}[!ht]
  \centering
  \includegraphics[width=\linewidth]{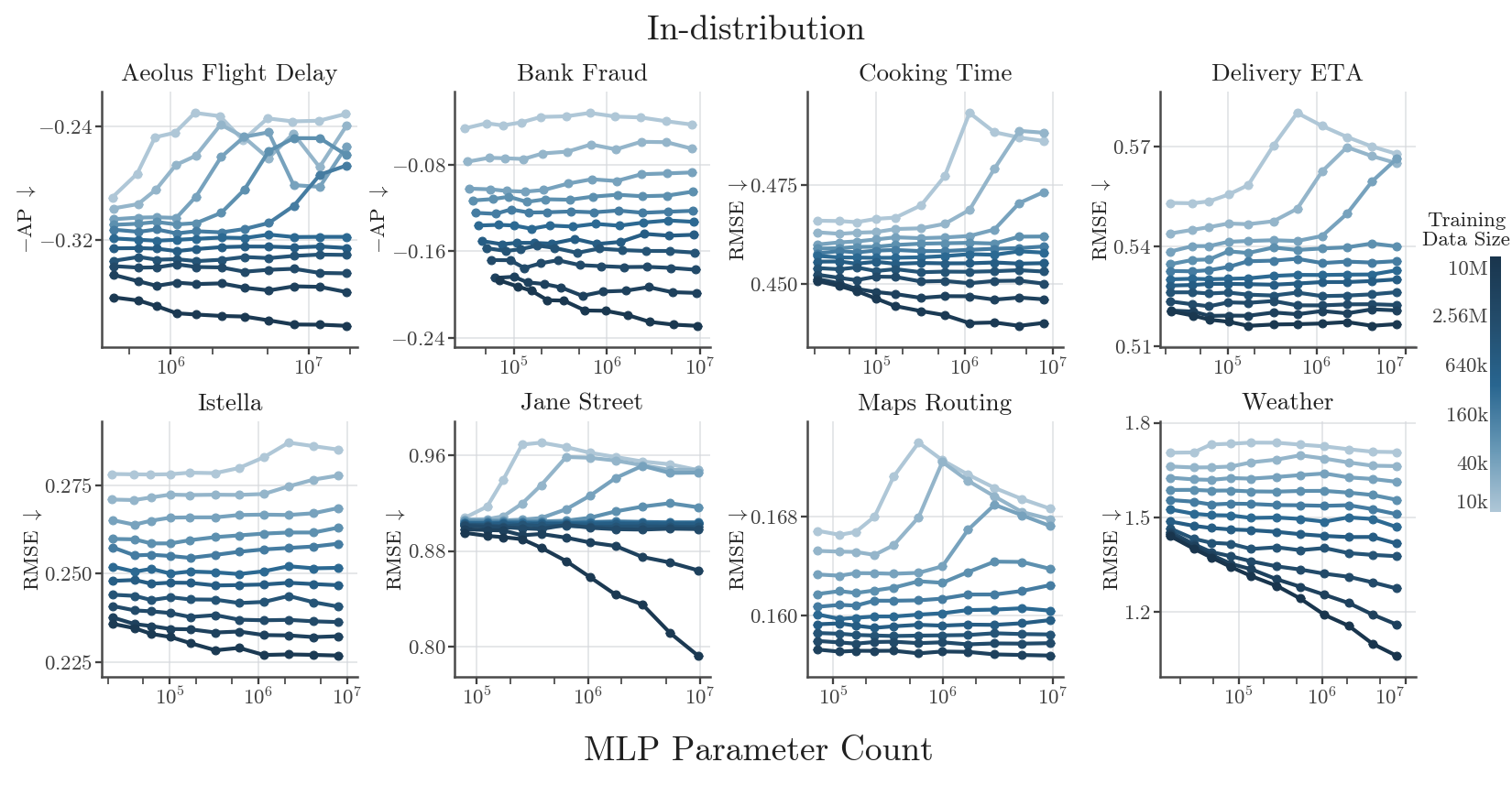}
  \caption{IID protocol: checkpoint and hyperparameter selection on IID
  validation, evaluation on IID test. Points show seed means; error bars show
  one standard deviation where multiple seeds are available.}
  \label{fig:full-sweep-iid}
\end{figure}

Figures~\ref{fig:full-sweep-default} and~\ref{fig:full-sweep-iid} show the
model-size sweeps on all eight datasets. For each training-set size and MLP
width, we select the learning rate, weight decay, and checkpoint using the
corresponding validation split. Darker curves indicate larger training sets.
The metrics are RMSE and negative average precision; lower is better.
Only completed runs are shown.

\FloatBarrier
\section{MLP and LightGBM with different validation splits}
\label{app:parametric-validation-splits}

\begin{figure}[!ht]
  \centering
  \includegraphics[width=\linewidth]{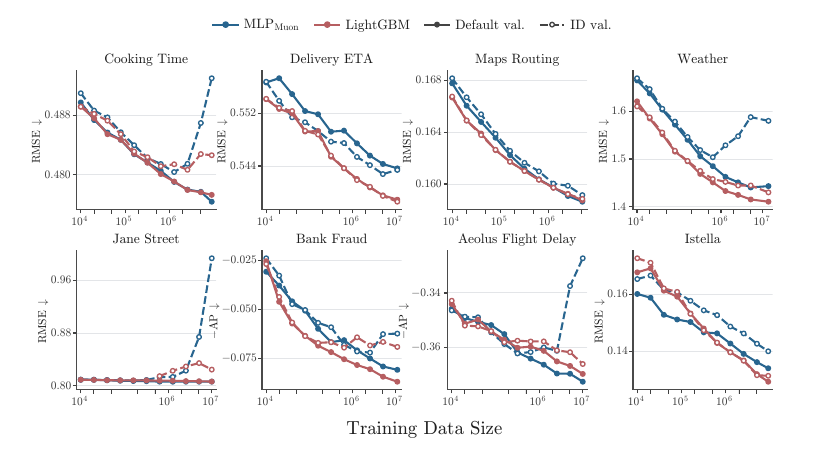}
  \caption{OOD test results: temporal test splits, or query for Istella.
  Blue curves show MLP with Muon; red curves show LightGBM.
  Solid curves with filled circles use default validation (temporal, or query-based
  for Istella); dashed curves with open circles use ID validation.}
  \label{fig:parametric-validation-ood}
\end{figure}

\begin{figure}[!ht]
  \centering
  \includegraphics[width=\linewidth]{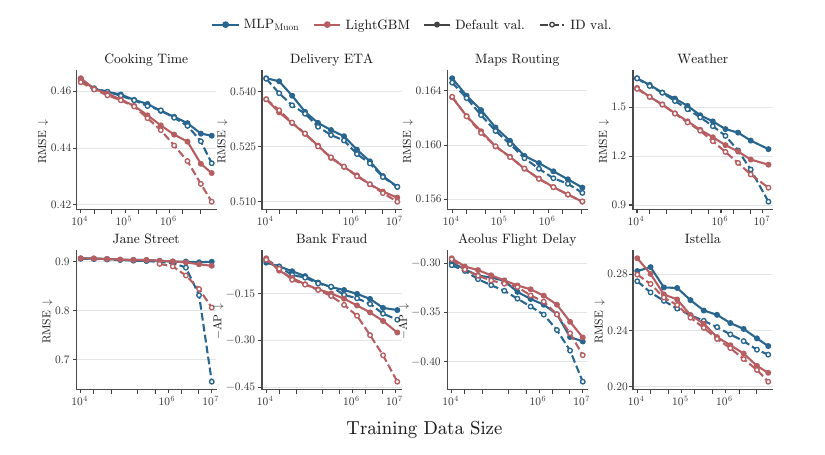}
  \caption{In-distribution test results: both validation protocols are
  evaluated on IID test. Colors and line styles follow
  Figure~\ref{fig:parametric-validation-ood}.}
  \label{fig:parametric-validation-iid}
\end{figure}

Figures~\ref{fig:parametric-validation-ood} and~\ref{fig:parametric-validation-iid}
compare MLP with Muon and LightGBM using default and ID validation on all eight
datasets. Each figure evaluates both validation protocols on the same test
split. Points show mean test scores. The metrics are RMSE
and negative average precision; lower is better.

\clearpage
\section{Full results for parametric methods}
\label{app:full-parametric-results}
Primary-test means and sample standard deviations.
\textemdash: no result. \textbf{Bold} marks first place and
\textit{italics} second place among multi-seed runs at each size
(sequential Welch tests, two-sided unadjusted $p<0.05$).
$\ddagger$: piecewise linear embeddings; $\dagger$: cosine feature embeddings.
``full'' is the full training split.
\par\begingroup
\setlength{\parskip}{0pt}
\par\addvspace{12pt}\noindent
\begin{minipage}{\linewidth}
\centering
{\normalsize\textbf{Cooking Time} / RMSE $\downarrow$\par}
\vspace{6pt}
\fontsize{8.5}{9.5}\selectfont
\setlength{\tabcolsep}{0.2pt}
\renewcommand{\arraystretch}{1.0}
\begin{tabular*}{\linewidth}{@{\extracolsep{\fill}}lccccccccccc@{}}
\toprule
 & \multicolumn{11}{c}{Training set size} \\
\cmidrule(lr){2-12}
Method & 10k & 20k & 40k & 80k & 160k & 320k & 640k & 1.28M & 2.56M & 5.12M & full \\
\midrule
\raisebox{-0.4\height}{\shortstack[l]{XGBoost}} & \raisebox{-0.4\height}{\shortstack{$0.4893$\\{\fontsize{6.5}{7.5}\selectfont\color[gray]{0.45}${\pm}0.0001$}}} & \raisebox{-0.4\height}{\shortstack{$0.4873$\\{\fontsize{6.5}{7.5}\selectfont\color[gray]{0.45}${\pm}0.0001$}}} & \raisebox{-0.4\height}{\shortstack{$0.4853$\\{\fontsize{6.5}{7.5}\selectfont\color[gray]{0.45}${\pm}0.0000$}}} & \raisebox{-0.4\height}{\shortstack{$0.4844$\\{\fontsize{6.5}{7.5}\selectfont\color[gray]{0.45}${\pm}0.0001$}}} & \raisebox{-0.4\height}{\shortstack{$0.4827$\\{\fontsize{6.5}{7.5}\selectfont\color[gray]{0.45}${\pm}0.0000$}}} & \raisebox{-0.4\height}{\shortstack{$0.4816$\\{\fontsize{6.5}{7.5}\selectfont\color[gray]{0.45}${\pm}0.0000$}}} & \raisebox{-0.4\height}{\shortstack{$0.4803$\\{\fontsize{6.5}{7.5}\selectfont\color[gray]{0.45}${\pm}0.0000$}}} & \raisebox{-0.4\height}{\shortstack{$0.4790$\\{\fontsize{6.5}{7.5}\selectfont\color[gray]{0.45}${\pm}0.0001$}}} & \raisebox{-0.4\height}{\shortstack{$0.4781$\\{\fontsize{6.5}{7.5}\selectfont\color[gray]{0.45}${\pm}0.0000$}}} & \raisebox{-0.4\height}{\shortstack{$0.4778$\\{\fontsize{6.5}{7.5}\selectfont\color[gray]{0.45}${\pm}0.0001$}}} & \raisebox{-0.4\height}{\shortstack{$0.4769$\\{\fontsize{6.5}{7.5}\selectfont\color[gray]{0.45}${\pm}0.0001$}}} \\
\bottomrule
\end{tabular*}
\end{minipage}\par
\par\addvspace{12pt}\noindent
\begin{minipage}{\linewidth}
\centering
{\normalsize\textbf{Delivery ETA} / RMSE $\downarrow$\par}
\vspace{6pt}
\fontsize{8.5}{9.5}\selectfont
\setlength{\tabcolsep}{0.2pt}
\renewcommand{\arraystretch}{1.0}
\begin{tabular*}{\linewidth}{@{\extracolsep{\fill}}lccccccccccc@{}}
\toprule
 & \multicolumn{11}{c}{Training set size} \\
\cmidrule(lr){2-12}
Method & 10k & 20k & 40k & 80k & 160k & 320k & 640k & 1.28M & 2.56M & 5.12M & full \\
\midrule
\raisebox{-0.4\height}{\shortstack[l]{XGBoost}} & \raisebox{-0.4\height}{\shortstack{$\mathit{0.5549}$\\{\fontsize{6.5}{7.5}\selectfont\color[gray]{0.45}${\pm}0.0001$}}} & \raisebox{-0.4\height}{\shortstack{$\mathit{0.5529}$\\{\fontsize{6.5}{7.5}\selectfont\color[gray]{0.45}${\pm}0.0001$}}} & \raisebox{-0.4\height}{\shortstack{$\mathit{0.5528}$\\{\fontsize{6.5}{7.5}\selectfont\color[gray]{0.45}${\pm}0.0001$}}} & \raisebox{-0.4\height}{\shortstack{$0.5503$\\{\fontsize{6.5}{7.5}\selectfont\color[gray]{0.45}${\pm}0.0001$}}} & \raisebox{-0.4\height}{\shortstack{$\mathbf{0.5491}$\\{\fontsize{6.5}{7.5}\selectfont\color[gray]{0.45}${\pm}0.0001$}}} & \raisebox{-0.4\height}{\shortstack{$\mathit{0.5458}$\\{\fontsize{6.5}{7.5}\selectfont\color[gray]{0.45}${\pm}0.0001$}}} & \raisebox{-0.4\height}{\shortstack{$\mathit{0.5442}$\\{\fontsize{6.5}{7.5}\selectfont\color[gray]{0.45}${\pm}0.0002$}}} & \raisebox{-0.4\height}{\shortstack{$\mathit{0.5422}$\\{\fontsize{6.5}{7.5}\selectfont\color[gray]{0.45}${\pm}0.0001$}}} & \raisebox{-0.4\height}{\shortstack{$\mathbf{0.5406}$\\{\fontsize{6.5}{7.5}\selectfont\color[gray]{0.45}${\pm}0.0001$}}} & \raisebox{-0.4\height}{\shortstack{$\mathit{0.5394}$\\{\fontsize{6.5}{7.5}\selectfont\color[gray]{0.45}${\pm}0.0001$}}} & \raisebox{-0.4\height}{\shortstack{$\mathbf{0.5380}$\\{\fontsize{6.5}{7.5}\selectfont\color[gray]{0.45}${\pm}0.0001$}}} \\
\bottomrule
\end{tabular*}
\end{minipage}\par
\par\addvspace{12pt}\noindent
\begin{minipage}{\linewidth}
\centering
{\normalsize\textbf{Maps Routing} / RMSE $\downarrow$\par}
\vspace{6pt}
\fontsize{8.5}{9.5}\selectfont
\setlength{\tabcolsep}{0.2pt}
\renewcommand{\arraystretch}{1.0}
\begin{tabular*}{\linewidth}{@{\extracolsep{\fill}}lcccccccccc@{}}
\toprule
 & \multicolumn{10}{c}{Training set size} \\
\cmidrule(lr){2-11}
Method & 10k & 20k & 40k & 80k & 160k & 320k & 640k & 1.28M & 2.56M & 5.12M \\
\midrule
\raisebox{-0.4\height}{\shortstack[l]{XGBoost}} & \raisebox{-0.4\height}{\shortstack{$0.1665$\\{\fontsize{6.5}{7.5}\selectfont\color[gray]{0.45}${\pm}0.0000$}}} & \raisebox{-0.4\height}{\shortstack{$0.1649$\\{\fontsize{6.5}{7.5}\selectfont\color[gray]{0.45}${\pm}0.0000$}}} & \raisebox{-0.4\height}{\shortstack{$0.1640$\\{\fontsize{6.5}{7.5}\selectfont\color[gray]{0.45}${\pm}0.0000$}}} & \raisebox{-0.4\height}{\shortstack{$0.1627$\\{\fontsize{6.5}{7.5}\selectfont\color[gray]{0.45}${\pm}0.0000$}}} & \raisebox{-0.4\height}{\shortstack{$0.1619$\\{\fontsize{6.5}{7.5}\selectfont\color[gray]{0.45}${\pm}0.0000$}}} & \raisebox{-0.4\height}{\shortstack{$0.1611$\\{\fontsize{6.5}{7.5}\selectfont\color[gray]{0.45}${\pm}0.0000$}}} & \raisebox{-0.4\height}{\shortstack{$0.1604$\\{\fontsize{6.5}{7.5}\selectfont\color[gray]{0.45}${\pm}0.0000$}}} & \raisebox{-0.4\height}{\shortstack{$0.1598$\\{\fontsize{6.5}{7.5}\selectfont\color[gray]{0.45}${\pm}0.0000$}}} & \raisebox{-0.4\height}{\shortstack{$0.1591$\\{\fontsize{6.5}{7.5}\selectfont\color[gray]{0.45}${\pm}0.0000$}}} & \raisebox{-0.4\height}{\shortstack{$0.1585$\\{\fontsize{6.5}{7.5}\selectfont\color[gray]{0.45}${\pm}0.0001$}}} \\
\bottomrule
\end{tabular*}
\end{minipage}\par
\par\addvspace{12pt}\noindent
\begin{minipage}{\linewidth}
\centering
{\normalsize\textbf{Weather} / RMSE $\downarrow$\par}
\vspace{6pt}
\fontsize{8.5}{9.5}\selectfont
\setlength{\tabcolsep}{0.2pt}
\renewcommand{\arraystretch}{1.0}
\begin{tabular*}{\linewidth}{@{\extracolsep{\fill}}lccccccccccc@{}}
\toprule
 & \multicolumn{11}{c}{Training set size} \\
\cmidrule(lr){2-12}
Method & 10k & 20k & 40k & 80k & 160k & 320k & 640k & 1.28M & 2.56M & 5.12M & full \\
\midrule
\raisebox{-0.4\height}{\shortstack[l]{XGBoost}} & \raisebox{-0.4\height}{\shortstack{$\mathit{1.6161}$\\{\fontsize{6.5}{7.5}\selectfont\color[gray]{0.45}${\pm}0.0006$}}} & \raisebox{-0.4\height}{\shortstack{$1.5875$\\{\fontsize{6.5}{7.5}\selectfont\color[gray]{0.45}${\pm}0.0007$}}} & \raisebox{-0.4\height}{\shortstack{$1.5582$\\{\fontsize{6.5}{7.5}\selectfont\color[gray]{0.45}${\pm}0.0006$}}} & \raisebox{-0.4\height}{\shortstack{$1.5232$\\{\fontsize{6.5}{7.5}\selectfont\color[gray]{0.45}${\pm}0.0008$}}} & \raisebox{-0.4\height}{\shortstack{$1.4984$\\{\fontsize{6.5}{7.5}\selectfont\color[gray]{0.45}${\pm}0.0009$}}} & \raisebox{-0.4\height}{\shortstack{$1.4716$\\{\fontsize{6.5}{7.5}\selectfont\color[gray]{0.45}${\pm}0.0006$}}} & \raisebox{-0.4\height}{\shortstack{$1.4466$\\{\fontsize{6.5}{7.5}\selectfont\color[gray]{0.45}${\pm}0.0005$}}} & \raisebox{-0.4\height}{\shortstack{$1.4318$\\{\fontsize{6.5}{7.5}\selectfont\color[gray]{0.45}${\pm}0.0004$}}} & \raisebox{-0.4\height}{\shortstack{$1.4221$\\{\fontsize{6.5}{7.5}\selectfont\color[gray]{0.45}${\pm}0.0004$}}} & \raisebox{-0.4\height}{\shortstack{$1.4136$\\{\fontsize{6.5}{7.5}\selectfont\color[gray]{0.45}${\pm}0.0014$}}} & \raisebox{-0.4\height}{\shortstack{$1.4073$\\{\fontsize{6.5}{7.5}\selectfont\color[gray]{0.45}${\pm}0.0007$}}} \\
\bottomrule
\end{tabular*}
\end{minipage}\par
\par\addvspace{12pt}\noindent
\begin{minipage}{\linewidth}
\centering
{\normalsize\textbf{Jane Street} / RMSE $\downarrow$\par}
\vspace{6pt}
\fontsize{8.5}{9.5}\selectfont
\setlength{\tabcolsep}{0.2pt}
\renewcommand{\arraystretch}{1.0}
\begin{tabular*}{\linewidth}{@{\extracolsep{\fill}}lccccccccccc@{}}
\toprule
 & \multicolumn{11}{c}{Training set size} \\
\cmidrule(lr){2-12}
Method & 10k & 20k & 40k & 80k & 160k & 320k & 640k & 1.28M & 2.56M & 5.12M & full \\
\midrule
\raisebox{-0.4\height}{\shortstack[l]{XGBoost}} & \raisebox{-0.4\height}{\shortstack{$0.8090$\\{\fontsize{6.5}{7.5}\selectfont\color[gray]{0.45}${\pm}0.0001$}}} & \raisebox{-0.4\height}{\shortstack{$\mathit{0.8087}$\\{\fontsize{6.5}{7.5}\selectfont\color[gray]{0.45}${\pm}0.0001$}}} & \raisebox{-0.4\height}{\shortstack{$0.8081$\\{\fontsize{6.5}{7.5}\selectfont\color[gray]{0.45}${\pm}0.0001$}}} & \raisebox{-0.4\height}{\shortstack{$0.8075$\\{\fontsize{6.5}{7.5}\selectfont\color[gray]{0.45}${\pm}0.0001$}}} & \raisebox{-0.4\height}{\shortstack{$0.8084$\\{\fontsize{6.5}{7.5}\selectfont\color[gray]{0.45}${\pm}0.0001$}}} & \raisebox{-0.4\height}{\shortstack{$0.8080$\\{\fontsize{6.5}{7.5}\selectfont\color[gray]{0.45}${\pm}0.0001$}}} & \raisebox{-0.4\height}{\shortstack{$0.8078$\\{\fontsize{6.5}{7.5}\selectfont\color[gray]{0.45}${\pm}0.0001$}}} & \raisebox{-0.4\height}{\shortstack{$0.8073$\\{\fontsize{6.5}{7.5}\selectfont\color[gray]{0.45}${\pm}0.0001$}}} & \raisebox{-0.4\height}{\shortstack{$0.8068$\\{\fontsize{6.5}{7.5}\selectfont\color[gray]{0.45}${\pm}0.0001$}}} & \raisebox{-0.4\height}{\shortstack{$0.8072$\\{\fontsize{6.5}{7.5}\selectfont\color[gray]{0.45}${\pm}0.0000$}}} & \raisebox{-0.4\height}{\shortstack{$0.8062$\\{\fontsize{6.5}{7.5}\selectfont\color[gray]{0.45}${\pm}0.0002$}}} \\
\bottomrule
\end{tabular*}
\end{minipage}\par
\par\addvspace{12pt}\noindent
\begin{minipage}{\linewidth}
\centering
{\normalsize\textbf{Bank Fraud Datafusion} / AP $\uparrow$\par}
\vspace{6pt}
\fontsize{8.5}{9.5}\selectfont
\setlength{\tabcolsep}{0.2pt}
\renewcommand{\arraystretch}{1.0}
\begin{tabular*}{\linewidth}{@{\extracolsep{\fill}}lccccccccccc@{}}
\toprule
 & \multicolumn{11}{c}{Training set size} \\
\cmidrule(lr){2-12}
Method & 10k & 20k & 40k & 80k & 160k & 320k & 640k & 1.28M & 2.56M & 5.12M & full \\
\midrule
\raisebox{-0.4\height}{\shortstack[l]{XGBoost}} & \raisebox{-0.4\height}{\shortstack{$0.0293$\\{\fontsize{6.5}{7.5}\selectfont\color[gray]{0.45}${\pm}0.0015$}}} & \raisebox{-0.4\height}{\shortstack{$0.0423$\\{\fontsize{6.5}{7.5}\selectfont\color[gray]{0.45}${\pm}0.0012$}}} & \raisebox{-0.4\height}{\shortstack{$0.0542$\\{\fontsize{6.5}{7.5}\selectfont\color[gray]{0.45}${\pm}0.0007$}}} & \raisebox{-0.4\height}{\shortstack{$0.0622$\\{\fontsize{6.5}{7.5}\selectfont\color[gray]{0.45}${\pm}0.0005$}}} & \raisebox{-0.4\height}{\shortstack{$0.0676$\\{\fontsize{6.5}{7.5}\selectfont\color[gray]{0.45}${\pm}0.0004$}}} & \raisebox{-0.4\height}{\shortstack{$0.0697$\\{\fontsize{6.5}{7.5}\selectfont\color[gray]{0.45}${\pm}0.0003$}}} & \raisebox{-0.4\height}{\shortstack{$0.0738$\\{\fontsize{6.5}{7.5}\selectfont\color[gray]{0.45}${\pm}0.0004$}}} & \raisebox{-0.4\height}{\shortstack{$0.0772$\\{\fontsize{6.5}{7.5}\selectfont\color[gray]{0.45}${\pm}0.0001$}}} & \raisebox{-0.4\height}{\shortstack{$0.0795$\\{\fontsize{6.5}{7.5}\selectfont\color[gray]{0.45}${\pm}0.0003$}}} & \raisebox{-0.4\height}{\shortstack{$0.0829$\\{\fontsize{6.5}{7.5}\selectfont\color[gray]{0.45}${\pm}0.0004$}}} & \raisebox{-0.4\height}{\shortstack{$\mathit{0.0849}$\\{\fontsize{6.5}{7.5}\selectfont\color[gray]{0.45}${\pm}0.0017$}}} \\
\bottomrule
\end{tabular*}
\end{minipage}\par
\par\addvspace{12pt}\noindent
\begin{minipage}{\linewidth}
\centering
{\normalsize\textbf{Aeolus Flight Delay} / AP $\uparrow$\par}
\vspace{6pt}
\fontsize{8.5}{9.5}\selectfont
\setlength{\tabcolsep}{0.2pt}
\renewcommand{\arraystretch}{1.0}
\begin{tabular*}{\linewidth}{@{\extracolsep{\fill}}lccccccccccc@{}}
\toprule
 & \multicolumn{11}{c}{Training set size} \\
\cmidrule(lr){2-12}
Method & 10k & 20k & 40k & 80k & 160k & 320k & 640k & 1.28M & 2.56M & 5.12M & full \\
\midrule
\raisebox{-0.4\height}{\shortstack[l]{XGBoost}} & \raisebox{-0.4\height}{\shortstack{$\mathit{0.3503}$\\{\fontsize{6.5}{7.5}\selectfont\color[gray]{0.45}${\pm}0.0005$}}} & \raisebox{-0.4\height}{\shortstack{$\mathbf{0.3551}$\\{\fontsize{6.5}{7.5}\selectfont\color[gray]{0.45}${\pm}0.0004$}}} & \raisebox{-0.4\height}{\shortstack{$\mathbf{0.3562}$\\{\fontsize{6.5}{7.5}\selectfont\color[gray]{0.45}${\pm}0.0003$}}} & \raisebox{-0.4\height}{\shortstack{$\mathbf{0.3581}$\\{\fontsize{6.5}{7.5}\selectfont\color[gray]{0.45}${\pm}0.0003$}}} & \raisebox{-0.4\height}{\shortstack{$\mathbf{0.3628}$\\{\fontsize{6.5}{7.5}\selectfont\color[gray]{0.45}${\pm}0.0005$}}} & \raisebox{-0.4\height}{\shortstack{$0.3630$\\{\fontsize{6.5}{7.5}\selectfont\color[gray]{0.45}${\pm}0.0003$}}} & \raisebox{-0.4\height}{\shortstack{$0.3656$\\{\fontsize{6.5}{7.5}\selectfont\color[gray]{0.45}${\pm}0.0002$}}} & \raisebox{-0.4\height}{\shortstack{$0.3655$\\{\fontsize{6.5}{7.5}\selectfont\color[gray]{0.45}${\pm}0.0002$}}} & \raisebox{-0.4\height}{\shortstack{$0.3648$\\{\fontsize{6.5}{7.5}\selectfont\color[gray]{0.45}${\pm}0.0002$}}} & \raisebox{-0.4\height}{\shortstack{$0.3689$\\{\fontsize{6.5}{7.5}\selectfont\color[gray]{0.45}${\pm}0.0002$}}} & \raisebox{-0.4\height}{\shortstack{$0.3712$\\{\fontsize{6.5}{7.5}\selectfont\color[gray]{0.45}${\pm}0.0005$}}} \\
\bottomrule
\end{tabular*}
\end{minipage}\par
\par\addvspace{12pt}\noindent
\begin{minipage}{\linewidth}
\centering
{\normalsize\textbf{Istella} / RMSE $\downarrow$\par}
\vspace{6pt}
\fontsize{8.5}{9.5}\selectfont
\setlength{\tabcolsep}{0.2pt}
\renewcommand{\arraystretch}{1.0}
\begin{tabular*}{\linewidth}{@{\extracolsep{\fill}}lccccccccccc@{}}
\toprule
 & \multicolumn{11}{c}{Training set size} \\
\cmidrule(lr){2-12}
Method & 10k & 20k & 40k & 80k & 160k & 320k & 640k & 1.28M & 2.56M & 5.12M & full \\
\midrule
\raisebox{-0.4\height}{\shortstack[l]{XGBoost}} & \raisebox{-0.4\height}{\shortstack{$0.1661$\\{\fontsize{6.5}{7.5}\selectfont\color[gray]{0.45}${\pm}0.0004$}}} & \raisebox{-0.4\height}{\shortstack{$0.1639$\\{\fontsize{6.5}{7.5}\selectfont\color[gray]{0.45}${\pm}0.0003$}}} & \raisebox{-0.4\height}{\shortstack{$0.1619$\\{\fontsize{6.5}{7.5}\selectfont\color[gray]{0.45}${\pm}0.0002$}}} & \raisebox{-0.4\height}{\shortstack{$0.1594$\\{\fontsize{6.5}{7.5}\selectfont\color[gray]{0.45}${\pm}0.0002$}}} & \raisebox{-0.4\height}{\shortstack{$0.1542$\\{\fontsize{6.5}{7.5}\selectfont\color[gray]{0.45}${\pm}0.0002$}}} & \raisebox{-0.4\height}{\shortstack{$0.1477$\\{\fontsize{6.5}{7.5}\selectfont\color[gray]{0.45}${\pm}0.0002$}}} & \raisebox{-0.4\height}{\shortstack{$\mathit{0.1427}$\\{\fontsize{6.5}{7.5}\selectfont\color[gray]{0.45}${\pm}0.0004$}}} & \raisebox{-0.4\height}{\shortstack{$0.1397$\\{\fontsize{6.5}{7.5}\selectfont\color[gray]{0.45}${\pm}0.0005$}}} & \raisebox{-0.4\height}{\shortstack{$0.1381$\\{\fontsize{6.5}{7.5}\selectfont\color[gray]{0.45}${\pm}0.0003$}}} & \raisebox{-0.4\height}{\shortstack{$0.1332$\\{\fontsize{6.5}{7.5}\selectfont\color[gray]{0.45}${\pm}0.0005$}}} & \raisebox{-0.4\height}{\shortstack{$0.1319$\\{\fontsize{6.5}{7.5}\selectfont\color[gray]{0.45}${\pm}0.0004$}}} \\
\bottomrule
\end{tabular*}
\end{minipage}\par
\endgroup

\clearpage
\section{Full results for foundation models}
\label{app:full-pfn-results}
Primary-test means and sample standard deviations.
\textemdash: no result. OOM: out of memory. \textbf{Bold} marks first place and
\textit{italics} second place among multi-seed runs at each size
(sequential Welch tests, two-sided unadjusted $p<0.05$).
Sizes without any foundation-model result are omitted.
\par\begingroup
\setlength{\parskip}{0pt}
\par\addvspace{12pt}\noindent
\begin{minipage}{\linewidth}
\centering
{\normalsize\textbf{Cooking Time} / RMSE $\downarrow$\par}
\vspace{6pt}
\fontsize{10}{11}\selectfont
\setlength{\tabcolsep}{0.2pt}
\renewcommand{\arraystretch}{1.0}
\begin{tabular*}{\linewidth}{@{\extracolsep{\fill}}lccccccc@{}}
\toprule
 & \multicolumn{7}{c}{Training set size} \\
\cmidrule(lr){2-8}
Method & 10k & 20k & 40k & 80k & 160k & 320k & 640k \\
\midrule
TabICL v2 & \raisebox{-0.4\height}{\shortstack{$0.4940$\\{\fontsize{8}{9}\selectfont\color[gray]{0.45}${\pm}0.0006$}}} & \raisebox{-0.4\height}{\shortstack{$0.4920$\\{\fontsize{8}{9}\selectfont\color[gray]{0.45}${\pm}0.0006$}}} & \raisebox{-0.4\height}{\shortstack{$0.4911$\\{\fontsize{8}{9}\selectfont\color[gray]{0.45}${\pm}0.0006$}}} & \raisebox{-0.4\height}{\shortstack{$0.4894$\\{\fontsize{8}{9}\selectfont\color[gray]{0.45}${\pm}0.0006$}}} & \raisebox{-0.4\height}{\shortstack{$0.4890$\\{\fontsize{8}{9}\selectfont\color[gray]{0.45}${\pm}0.0006$}}} & \raisebox{-0.4\height}{\shortstack{$0.4883$\\{\fontsize{8}{9}\selectfont\color[gray]{0.45}${\pm}0.0006$}}} & \raisebox{-0.4\height}{\shortstack{$0.4881$\\{\fontsize{8}{9}\selectfont\color[gray]{0.45}${\pm}0.0006$}}} \\[7pt]
TabPFN v3 & \raisebox{-0.4\height}{\shortstack{$0.4917$\\{\fontsize{8}{9}\selectfont\color[gray]{0.45}${\pm}0.0005$}}} & \raisebox{-0.4\height}{\shortstack{$0.4897$\\{\fontsize{8}{9}\selectfont\color[gray]{0.45}${\pm}0.0002$}}} & \raisebox{-0.4\height}{\shortstack{$0.4886$\\{\fontsize{8}{9}\selectfont\color[gray]{0.45}${\pm}0.0002$}}} & \raisebox{-0.4\height}{\shortstack{$0.4878$\\{\fontsize{8}{9}\selectfont\color[gray]{0.45}${\pm}0.0003$}}} & \raisebox{-0.4\height}{\shortstack{$0.4873$\\{\fontsize{8}{9}\selectfont\color[gray]{0.45}${\pm}0.0002$}}} & \raisebox{-0.4\height}{\shortstack{$0.4872$\\{\fontsize{8}{9}\selectfont\color[gray]{0.45}${\pm}0.0003$}}} & \raisebox{-0.4\height}{\shortstack{$0.4870$\\{\fontsize{8}{9}\selectfont\color[gray]{0.45}${\pm}0.0004$}}} \\
\bottomrule
\end{tabular*}
\end{minipage}\par
\par\addvspace{12pt}\noindent
\begin{minipage}{\linewidth}
\centering
{\normalsize\textbf{Delivery ETA} / RMSE $\downarrow$\par}
\vspace{6pt}
\fontsize{10}{11}\selectfont
\setlength{\tabcolsep}{0.2pt}
\renewcommand{\arraystretch}{1.0}
\begin{tabular*}{\linewidth}{@{\extracolsep{\fill}}lccccccc@{}}
\toprule
 & \multicolumn{7}{c}{Training set size} \\
\cmidrule(lr){2-8}
Method & 10k & 20k & 40k & 80k & 160k & 320k & 640k \\
\midrule
TabICL v2 & \raisebox{-0.4\height}{\shortstack{$0.5548$\\{\fontsize{8}{9}\selectfont\color[gray]{0.45}${\pm}0.0005$}}} & \raisebox{-0.4\height}{\shortstack{$0.5554$\\{\fontsize{8}{9}\selectfont\color[gray]{0.45}${\pm}0.0003$}}} & \raisebox{-0.4\height}{\shortstack{$0.5555$\\{\fontsize{8}{9}\selectfont\color[gray]{0.45}${\pm}0.0004$}}} & \raisebox{-0.4\height}{\shortstack{$0.5543$\\{\fontsize{8}{9}\selectfont\color[gray]{0.45}${\pm}0.0004$}}} & \raisebox{-0.4\height}{\shortstack{$0.5553$\\{\fontsize{8}{9}\selectfont\color[gray]{0.45}${\pm}0.0006$}}} & \raisebox{-0.4\height}{\shortstack{$0.5544$\\{\fontsize{8}{9}\selectfont\color[gray]{0.45}${\pm}0.0006$}}} & \raisebox{-0.4\height}{\shortstack{$0.5552$\\{\fontsize{8}{9}\selectfont\color[gray]{0.45}${\pm}0.0008$}}} \\[7pt]
TabPFN v3 & \raisebox{-0.4\height}{\shortstack{$0.5552$\\{\fontsize{8}{9}\selectfont\color[gray]{0.45}${\pm}0.0005$}}} & \raisebox{-0.4\height}{\shortstack{$0.5546$\\{\fontsize{8}{9}\selectfont\color[gray]{0.45}${\pm}0.0004$}}} & \raisebox{-0.4\height}{\shortstack{$0.5534$\\{\fontsize{8}{9}\selectfont\color[gray]{0.45}${\pm}0.0004$}}} & \raisebox{-0.4\height}{\shortstack{$0.5526$\\{\fontsize{8}{9}\selectfont\color[gray]{0.45}${\pm}0.0002$}}} & \raisebox{-0.4\height}{\shortstack{$0.5531$\\{\fontsize{8}{9}\selectfont\color[gray]{0.45}${\pm}0.0004$}}} & \raisebox{-0.4\height}{\shortstack{$0.5528$\\{\fontsize{8}{9}\selectfont\color[gray]{0.45}${\pm}0.0004$}}} & \raisebox{-0.4\height}{\shortstack{$0.5542$\\{\fontsize{8}{9}\selectfont\color[gray]{0.45}${\pm}0.0008$}}} \\
\bottomrule
\end{tabular*}
\end{minipage}\par
\par\addvspace{12pt}\noindent
\begin{minipage}{\linewidth}
\centering
{\normalsize\textbf{Maps Routing} / RMSE $\downarrow$\par}
\vspace{6pt}
\fontsize{10}{11}\selectfont
\setlength{\tabcolsep}{0.2pt}
\renewcommand{\arraystretch}{1.0}
\begin{tabular*}{\linewidth}{@{\extracolsep{\fill}}lccccccc@{}}
\toprule
 & \multicolumn{7}{c}{Training set size} \\
\cmidrule(lr){2-8}
Method & 10k & 20k & 40k & 80k & 160k & 320k & 640k \\
\midrule
TabICL v2 & \raisebox{-0.4\height}{\shortstack{$0.1680$\\{\fontsize{8}{9}\selectfont\color[gray]{0.45}${\pm}0.0003$}}} & \raisebox{-0.4\height}{\shortstack{$0.1670$\\{\fontsize{8}{9}\selectfont\color[gray]{0.45}${\pm}0.0003$}}} & \raisebox{-0.4\height}{\shortstack{$0.1670$\\{\fontsize{8}{9}\selectfont\color[gray]{0.45}${\pm}0.0003$}}} & \raisebox{-0.4\height}{\shortstack{$0.1663$\\{\fontsize{8}{9}\selectfont\color[gray]{0.45}${\pm}0.0003$}}} & \raisebox{-0.4\height}{\shortstack{$0.1662$\\{\fontsize{8}{9}\selectfont\color[gray]{0.45}${\pm}0.0003$}}} & \raisebox{-0.4\height}{\shortstack{$0.1663$\\{\fontsize{8}{9}\selectfont\color[gray]{0.45}${\pm}0.0003$}}} & \raisebox{-0.4\height}{\shortstack{$0.1664$\\{\fontsize{8}{9}\selectfont\color[gray]{0.45}${\pm}0.0003$}}} \\[7pt]
TabPFN v3 & \raisebox{-0.4\height}{\shortstack{$0.1672$\\{\fontsize{8}{9}\selectfont\color[gray]{0.45}${\pm}0.0002$}}} & \raisebox{-0.4\height}{\shortstack{$0.1661$\\{\fontsize{8}{9}\selectfont\color[gray]{0.45}${\pm}0.0001$}}} & \raisebox{-0.4\height}{\shortstack{$0.1660$\\{\fontsize{8}{9}\selectfont\color[gray]{0.45}${\pm}0.0001$}}} & \raisebox{-0.4\height}{\shortstack{$0.1653$\\{\fontsize{8}{9}\selectfont\color[gray]{0.45}${\pm}0.0001$}}} & \raisebox{-0.4\height}{\shortstack{$0.1650$\\{\fontsize{8}{9}\selectfont\color[gray]{0.45}${\pm}0.0001$}}} & \raisebox{-0.4\height}{\shortstack{$0.1649$\\{\fontsize{8}{9}\selectfont\color[gray]{0.45}${\pm}0.0001$}}} & \raisebox{-0.4\height}{\shortstack{$0.1634$\\{\fontsize{8}{9}\selectfont\color[gray]{0.45}${\pm}0.0008$}}} \\
\bottomrule
\end{tabular*}
\end{minipage}\par
\par\addvspace{12pt}\noindent
\begin{minipage}{\linewidth}
\centering
{\normalsize\textbf{Weather} / RMSE $\downarrow$\par}
\vspace{6pt}
\fontsize{10}{11}\selectfont
\setlength{\tabcolsep}{0.2pt}
\renewcommand{\arraystretch}{1.0}
\begin{tabular*}{\linewidth}{@{\extracolsep{\fill}}lccccccc@{}}
\toprule
 & \multicolumn{7}{c}{Training set size} \\
\cmidrule(lr){2-8}
Method & 10k & 20k & 40k & 80k & 160k & 320k & 640k \\
\midrule
TabICL v2 & \raisebox{-0.4\height}{\shortstack{$1.6205$\\{\fontsize{8}{9}\selectfont\color[gray]{0.45}${\pm}0.0042$}}} & \raisebox{-0.4\height}{\shortstack{$1.6064$\\{\fontsize{8}{9}\selectfont\color[gray]{0.45}${\pm}0.0039$}}} & \raisebox{-0.4\height}{\shortstack{$1.5815$\\{\fontsize{8}{9}\selectfont\color[gray]{0.45}${\pm}0.0039$}}} & \raisebox{-0.4\height}{\shortstack{$1.5530$\\{\fontsize{8}{9}\selectfont\color[gray]{0.45}${\pm}0.0036$}}} & \raisebox{-0.4\height}{\shortstack{$1.5252$\\{\fontsize{8}{9}\selectfont\color[gray]{0.45}${\pm}0.0035$}}} & \raisebox{-0.4\height}{\shortstack{$1.5000$\\{\fontsize{8}{9}\selectfont\color[gray]{0.45}${\pm}0.0032$}}} & \raisebox{-0.4\height}{\shortstack{$1.4797$\\{\fontsize{8}{9}\selectfont\color[gray]{0.45}${\pm}0.0033$}}} \\[7pt]
TabPFN v3 & \raisebox{-0.4\height}{\shortstack{$1.5968$\\{\fontsize{8}{9}\selectfont\color[gray]{0.45}${\pm}0.0021$}}} & \raisebox{-0.4\height}{\shortstack{$1.5785$\\{\fontsize{8}{9}\selectfont\color[gray]{0.45}${\pm}0.0023$}}} & \raisebox{-0.4\height}{\shortstack{$1.5550$\\{\fontsize{8}{9}\selectfont\color[gray]{0.45}${\pm}0.0020$}}} & \raisebox{-0.4\height}{\shortstack{$1.5266$\\{\fontsize{8}{9}\selectfont\color[gray]{0.45}${\pm}0.0024$}}} & \raisebox{-0.4\height}{\shortstack{$1.5030$\\{\fontsize{8}{9}\selectfont\color[gray]{0.45}${\pm}0.0016$}}} & \raisebox{-0.4\height}{\shortstack{$1.4831$\\{\fontsize{8}{9}\selectfont\color[gray]{0.45}${\pm}0.0017$}}} & \raisebox{-0.4\height}{\shortstack{$1.4501$\\{\fontsize{8}{9}\selectfont\color[gray]{0.45}${\pm}0.0022$}}} \\
\bottomrule
\end{tabular*}
\end{minipage}\par
\par\addvspace{12pt}\noindent
\begin{minipage}{\linewidth}
\centering
{\normalsize\textbf{Jane Street} / RMSE $\downarrow$\par}
\vspace{6pt}
\fontsize{10}{11}\selectfont
\setlength{\tabcolsep}{0.2pt}
\renewcommand{\arraystretch}{1.0}
\begin{tabular*}{\linewidth}{@{\extracolsep{\fill}}lccccccc@{}}
\toprule
 & \multicolumn{7}{c}{Training set size} \\
\cmidrule(lr){2-8}
Method & 10k & 20k & 40k & 80k & 160k & 320k & 640k \\
\midrule
TabICL v2 & \raisebox{-0.4\height}{\shortstack{$0.8104$\\{\fontsize{8}{9}\selectfont\color[gray]{0.45}${\pm}0.0002$}}} & \raisebox{-0.4\height}{\shortstack{$0.8091$\\{\fontsize{8}{9}\selectfont\color[gray]{0.45}${\pm}0.0001$}}} & \raisebox{-0.4\height}{\shortstack{$0.8082$\\{\fontsize{8}{9}\selectfont\color[gray]{0.45}${\pm}0.0001$}}} & \raisebox{-0.4\height}{\shortstack{$0.8081$\\{\fontsize{8}{9}\selectfont\color[gray]{0.45}${\pm}0.0002$}}} & \raisebox{-0.4\height}{\shortstack{$0.8090$\\{\fontsize{8}{9}\selectfont\color[gray]{0.45}${\pm}0.0003$}}} & \raisebox{-0.4\height}{\shortstack{$0.8099$\\{\fontsize{8}{9}\selectfont\color[gray]{0.45}${\pm}0.0004$}}} & \raisebox{-0.4\height}{\shortstack{$0.8115$\\{\fontsize{8}{9}\selectfont\color[gray]{0.45}${\pm}0.0004$}}} \\[7pt]
TabPFN v3 & \raisebox{-0.4\height}{\shortstack{$\mathbf{0.8088}$\\{\fontsize{8}{9}\selectfont\color[gray]{0.45}${\pm}0.0000$}}} & \raisebox{-0.4\height}{\shortstack{$\mathbf{0.8084}$\\{\fontsize{8}{9}\selectfont\color[gray]{0.45}${\pm}0.0000$}}} & \raisebox{-0.4\height}{\shortstack{$0.8078$\\{\fontsize{8}{9}\selectfont\color[gray]{0.45}${\pm}0.0001$}}} & \raisebox{-0.4\height}{\shortstack{$0.8074$\\{\fontsize{8}{9}\selectfont\color[gray]{0.45}${\pm}0.0001$}}} & \raisebox{-0.4\height}{\shortstack{$0.8075$\\{\fontsize{8}{9}\selectfont\color[gray]{0.45}${\pm}0.0001$}}} & \raisebox{-0.4\height}{\shortstack{$0.8072$\\{\fontsize{8}{9}\selectfont\color[gray]{0.45}${\pm}0.0001$}}} & \raisebox{-0.4\height}{\shortstack{$\mathbf{0.8061}$\\{\fontsize{8}{9}\selectfont\color[gray]{0.45}${\pm}0.0001$}}} \\
\bottomrule
\end{tabular*}
\end{minipage}\par
\par\addvspace{12pt}\noindent
\begin{minipage}{\linewidth}
\centering
{\normalsize\textbf{Bank Fraud Datafusion} / AP $\uparrow$\par}
\vspace{6pt}
\fontsize{10}{11}\selectfont
\setlength{\tabcolsep}{0.2pt}
\renewcommand{\arraystretch}{1.0}
\begin{tabular*}{\linewidth}{@{\extracolsep{\fill}}lccccccc@{}}
\toprule
 & \multicolumn{7}{c}{Training set size} \\
\cmidrule(lr){2-8}
Method & 10k & 20k & 40k & 80k & 160k & 320k & 640k \\
\midrule
TabICL v2 & \raisebox{-0.4\height}{\shortstack{$0.0263$\\{\fontsize{8}{9}\selectfont\color[gray]{0.45}${\pm}0.0014$}}} & \raisebox{-0.4\height}{\shortstack{$0.0421$\\{\fontsize{8}{9}\selectfont\color[gray]{0.45}${\pm}0.0006$}}} & \raisebox{-0.4\height}{\shortstack{$0.0523$\\{\fontsize{8}{9}\selectfont\color[gray]{0.45}${\pm}0.0007$}}} & \raisebox{-0.4\height}{\shortstack{$0.0542$\\{\fontsize{8}{9}\selectfont\color[gray]{0.45}${\pm}0.0005$}}} & \raisebox{-0.4\height}{\shortstack{$0.0556$\\{\fontsize{8}{9}\selectfont\color[gray]{0.45}${\pm}0.0004$}}} & \raisebox{-0.4\height}{\shortstack{$0.0583$\\{\fontsize{8}{9}\selectfont\color[gray]{0.45}${\pm}0.0006$}}} & \raisebox{-0.4\height}{\shortstack{$0.0596$\\{\fontsize{8}{9}\selectfont\color[gray]{0.45}${\pm}0.0006$}}} \\[7pt]
TabPFN v3 & \raisebox{-0.4\height}{\shortstack{$0.0257$\\{\fontsize{8}{9}\selectfont\color[gray]{0.45}${\pm}0.0017$}}} & \raisebox{-0.4\height}{\shortstack{$0.0443$\\{\fontsize{8}{9}\selectfont\color[gray]{0.45}${\pm}0.0013$}}} & \raisebox{-0.4\height}{\shortstack{$0.0528$\\{\fontsize{8}{9}\selectfont\color[gray]{0.45}${\pm}0.0012$}}} & \raisebox{-0.4\height}{\shortstack{$0.0573$\\{\fontsize{8}{9}\selectfont\color[gray]{0.45}${\pm}0.0005$}}} & \raisebox{-0.4\height}{\shortstack{$0.0612$\\{\fontsize{8}{9}\selectfont\color[gray]{0.45}${\pm}0.0019$}}} & \raisebox{-0.4\height}{\shortstack{$0.0659$\\{\fontsize{8}{9}\selectfont\color[gray]{0.45}${\pm}0.0016$}}} & \raisebox{-0.4\height}{\shortstack{$0.0667$\\{\fontsize{8}{9}\selectfont\color[gray]{0.45}${\pm}0.0017$}}} \\
\bottomrule
\end{tabular*}
\end{minipage}\par
\par\addvspace{12pt}\noindent
\begin{minipage}{\linewidth}
\centering
{\normalsize\textbf{Aeolus Flight Delay} / AP $\uparrow$\par}
\vspace{6pt}
\fontsize{10}{11}\selectfont
\setlength{\tabcolsep}{0.2pt}
\renewcommand{\arraystretch}{1.0}
\begin{tabular*}{\linewidth}{@{\extracolsep{\fill}}lccccccc@{}}
\toprule
 & \multicolumn{7}{c}{Training set size} \\
\cmidrule(lr){2-8}
Method & 10k & 20k & 40k & 80k & 160k & 320k & 640k \\
\midrule
TabICL v2 & \raisebox{-0.4\height}{\shortstack{$0.3486$\\{\fontsize{8}{9}\selectfont\color[gray]{0.45}${\pm}0.0014$}}} & \raisebox{-0.4\height}{\shortstack{$0.3525$\\{\fontsize{8}{9}\selectfont\color[gray]{0.45}${\pm}0.0010$}}} & \raisebox{-0.4\height}{\shortstack{$0.3523$\\{\fontsize{8}{9}\selectfont\color[gray]{0.45}${\pm}0.0015$}}} & \raisebox{-0.4\height}{\shortstack{$0.3545$\\{\fontsize{8}{9}\selectfont\color[gray]{0.45}${\pm}0.0018$}}} & \raisebox{-0.4\height}{\shortstack{$0.3566$\\{\fontsize{8}{9}\selectfont\color[gray]{0.45}${\pm}0.0022$}}} & \raisebox{-0.4\height}{\shortstack{$0.3571$\\{\fontsize{8}{9}\selectfont\color[gray]{0.45}${\pm}0.0026$}}} & \raisebox{-0.4\height}{\shortstack{$0.3559$\\{\fontsize{8}{9}\selectfont\color[gray]{0.45}${\pm}0.0035$}}} \\[7pt]
TabPFN v3 & \raisebox{-0.4\height}{\shortstack{$0.3541$\\{\fontsize{8}{9}\selectfont\color[gray]{0.45}${\pm}0.0003$}}} & \raisebox{-0.4\height}{\shortstack{$0.3582$\\{\fontsize{8}{9}\selectfont\color[gray]{0.45}${\pm}0.0004$}}} & \raisebox{-0.4\height}{\shortstack{$0.3576$\\{\fontsize{8}{9}\selectfont\color[gray]{0.45}${\pm}0.0005$}}} & \raisebox{-0.4\height}{\shortstack{$0.3598$\\{\fontsize{8}{9}\selectfont\color[gray]{0.45}${\pm}0.0006$}}} & \raisebox{-0.4\height}{\shortstack{$0.3626$\\{\fontsize{8}{9}\selectfont\color[gray]{0.45}${\pm}0.0008$}}} & \raisebox{-0.4\height}{\shortstack{$\mathit{0.3642}$\\{\fontsize{8}{9}\selectfont\color[gray]{0.45}${\pm}0.0008$}}} & \raisebox{-0.4\height}{\shortstack{$0.3640$\\{\fontsize{8}{9}\selectfont\color[gray]{0.45}${\pm}0.0008$}}} \\
\bottomrule
\end{tabular*}
\end{minipage}\par
\par\addvspace{12pt}\noindent
\begin{minipage}{\linewidth}
\centering
{\normalsize\textbf{Istella} / RMSE $\downarrow$\par}
\vspace{6pt}
\fontsize{10}{11}\selectfont
\setlength{\tabcolsep}{0.2pt}
\renewcommand{\arraystretch}{1.0}
\begin{tabular*}{\linewidth}{@{\extracolsep{\fill}}lccccccc@{}}
\toprule
 & \multicolumn{7}{c}{Training set size} \\
\cmidrule(lr){2-8}
Method & 10k & 20k & 40k & 80k & 160k & 320k & 640k \\
\midrule
TabICL v2 & \raisebox{-0.4\height}{\shortstack{$0.1655$\\{\fontsize{8}{9}\selectfont\color[gray]{0.45}${\pm}0.0007$}}} & \raisebox{-0.4\height}{\shortstack{$0.1654$\\{\fontsize{8}{9}\selectfont\color[gray]{0.45}${\pm}0.0009$}}} & \raisebox{-0.4\height}{\shortstack{$0.1610$\\{\fontsize{8}{9}\selectfont\color[gray]{0.45}${\pm}0.0006$}}} & \raisebox{-0.4\height}{\shortstack{$0.1601$\\{\fontsize{8}{9}\selectfont\color[gray]{0.45}${\pm}0.0006$}}} & \raisebox{-0.4\height}{\shortstack{$0.1593$\\{\fontsize{8}{9}\selectfont\color[gray]{0.45}${\pm}0.0009$}}} & \raisebox{-0.4\height}{\shortstack{$0.1584$\\{\fontsize{8}{9}\selectfont\color[gray]{0.45}${\pm}0.0011$}}} & \raisebox{-0.4\height}{\shortstack{$0.1594$\\{\fontsize{8}{9}\selectfont\color[gray]{0.45}${\pm}0.0013$}}} \\[7pt]
TabPFN v3 & \raisebox{-0.4\height}{\shortstack{$0.1630$\\{\fontsize{8}{9}\selectfont\color[gray]{0.45}${\pm}0.0005$}}} & \raisebox{-0.4\height}{\shortstack{$0.1622$\\{\fontsize{8}{9}\selectfont\color[gray]{0.45}${\pm}0.0005$}}} & \raisebox{-0.4\height}{\shortstack{$0.1630$\\{\fontsize{8}{9}\selectfont\color[gray]{0.45}${\pm}0.0005$}}} & \raisebox{-0.4\height}{\shortstack{$0.1634$\\{\fontsize{8}{9}\selectfont\color[gray]{0.45}${\pm}0.0004$}}} & \raisebox{-0.4\height}{\shortstack{$0.1646$\\{\fontsize{8}{9}\selectfont\color[gray]{0.45}${\pm}0.0002$}}} & \raisebox{-0.4\height}{\shortstack{$0.1636$\\{\fontsize{8}{9}\selectfont\color[gray]{0.45}${\pm}0.0028$}}} & \raisebox{-0.4\height}{\shortstack{$0.1603$\\{\fontsize{8}{9}\selectfont\color[gray]{0.45}${\pm}0.0003$}}} \\
\bottomrule
\end{tabular*}
\end{minipage}\par
\endgroup

\end{document}